\documentclass{article}

\usepackage[preprint]{neurips_2025}

\usepackage[utf8]{inputenc} 
\usepackage[T1]{fontenc}    
\usepackage{hyperref}       
\usepackage{url}            
\usepackage{booktabs}       
\usepackage{amsfonts}       
\usepackage{nicefrac}       
\usepackage{microtype}      
\usepackage{xcolor}         
\usepackage{soul}
\usepackage{amssymb}
\usepackage{amsmath}
\usepackage{natbib}
\usepackage{multirow}
\usepackage{graphicx}
\usepackage[normalem]{ulem}
\usepackage{wrapfig}
\usepackage{adjustbox}
\usepackage{bm}
\usepackage{enumitem}
\usepackage{colortbl}
\usepackage{comment}
\newcommand{\insertimagescontextirseverevtwo}[1]{%
    \includegraphics[width=0.2\linewidth]{contextir-severe-v2/#1-lq.jpg} &
    \includegraphics[width=0.2\linewidth]{contextir-severe-v2/#1-ref1.jpg} &
    \includegraphics[width=0.2\linewidth]{contextir-severe-v2/#1-codeformer.jpg} &
    \includegraphics[width=0.2\linewidth]{contextir-severe-v2/#1-diffbir.jpg} &
    \includegraphics[width=0.2\linewidth]{contextir-severe-v2/#1-refldm.jpg} &
    \includegraphics[width=0.2\linewidth]{contextir-severe-v2/#1-restorerid.jpg} &
    \includegraphics[width=0.2\linewidth]{contextir-severe-v2/#1-ours.jpg} &
    \includegraphics[width=0.2\linewidth]{contextir-severe-v2/#1-hq.jpg} %
}

\newcommand{\insertimagescontextirmoderatevtwo}[1]{%
    \includegraphics[width=0.2\linewidth]{contextir-moderate-v2/#1-lq.jpg} &
    \includegraphics[width=0.2\linewidth]{contextir-moderate-v2/#1-ref1.jpg} &
    \includegraphics[width=0.2\linewidth]{contextir-moderate-v2/#1-codeformer.jpg} &
    \includegraphics[width=0.2\linewidth]{contextir-moderate-v2/#1-diffbir.jpg} &
    \includegraphics[width=0.2\linewidth]{contextir-moderate-v2/#1-refldm.jpg} &
    \includegraphics[width=0.2\linewidth]{contextir-moderate-v2/#1-restorerid.jpg} &
    \includegraphics[width=0.2\linewidth]{contextir-moderate-v2/#1-ours.jpg} &
    \includegraphics[width=0.2\linewidth]{contextir-moderate-v2/#1-hq.jpg} %
}

\newcommand{\suppfigsevere}[1]{%
    \includegraphics[width=0.2\linewidth]{supp-severe/#1-lq.jpg} &
    \includegraphics[width=0.2\linewidth]{supp-severe/#1-ref1.jpg} &
    \includegraphics[width=0.2\linewidth]{supp-severe/#1-codeformer.jpg} &
    \includegraphics[width=0.2\linewidth]{supp-severe/#1-diffbir.jpg} &
    \includegraphics[width=0.2\linewidth]{supp-severe/#1-refldm.jpg} &
    \includegraphics[width=0.2\linewidth]{supp-severe/#1-restorerid.jpg} &
    \includegraphics[width=0.2\linewidth]{supp-severe/#1-ours.jpg} &
    \includegraphics[width=0.2\linewidth]{supp-severe/#1-hq.jpg} %
}

\newcommand{\suppfigmoderate}[1]{%
    \includegraphics[width=0.2\linewidth]{supp-moderate/#1-lq.jpg} &
    \includegraphics[width=0.2\linewidth]{supp-moderate/#1-ref1.jpg} &
    \includegraphics[width=0.2\linewidth]{supp-moderate/#1-codeformer.jpg} &
    \includegraphics[width=0.2\linewidth]{supp-moderate/#1-diffbir.jpg} &
    \includegraphics[width=0.2\linewidth]{supp-moderate/#1-refldm.jpg} &
    \includegraphics[width=0.2\linewidth]{supp-moderate/#1-restorerid.jpg} &
    \includegraphics[width=0.2\linewidth]{supp-moderate/#1-ours.jpg} &
    \includegraphics[width=0.2\linewidth]{supp-moderate/#1-hq.jpg} %
}

\newcommand{\suppfigceleba}[1]{%
    \includegraphics[width=0.2\linewidth]{supp-celeba/#1-lq.jpg} &
    \includegraphics[width=0.2\linewidth]{supp-celeba/#1-ref1.jpg} &
    \includegraphics[width=0.2\linewidth]{supp-celeba/#1-codeformer.jpg} &
    \includegraphics[width=0.2\linewidth]{supp-celeba/#1-diffbir.jpg} &
    \includegraphics[width=0.2\linewidth]{supp-celeba/#1-refldm.jpg} &
    \includegraphics[width=0.2\linewidth]{supp-celeba/#1-restorerid.jpg} &
    \includegraphics[width=0.2\linewidth]{supp-celeba/#1-ours.jpg} &
    \includegraphics[width=0.2\linewidth]{supp-celeba/#1-hq.jpg} %
}

\title{Reference-Guided Identity Preserving\\ Face Restoration}

\author{%
Mo Zhou$^{1,2}$\thanks{Work done during internship at Google LLC.}
\quad
Keren Ye$^1$
\quad
Viraj Shah$^1$
\quad
Kangfu Mei$^1$
\quad
Mauricio Delbracio$^1$
\\
{\bf Peyman Milanfar}$^1$
\quad
{\bf Vishal M.~Patel}$^2$
\quad
{\bf Hossein Talebi}$^1$
\\
$^1$Google \quad $^2$Johns Hopkins University
}

\begin{document}

\maketitle

\begin{abstract}

%
%
%
%
%
%
%
%
%

Preserving face identity is a critical yet persistent challenge in diffusion-based image restoration.
While reference faces offer a path forward, existing reference-based methods often fail to fully exploit their potential.
This paper introduces a novel approach that maximizes reference face utility for improved face restoration and identity preservation.
Our method makes three key contributions:
1) Composite Context, a comprehensive representation that fuses multi-level (high- and low-level) information from the reference face, offering richer guidance than prior singular representations.
2) Hard Example Identity Loss, a novel loss function that leverages the reference face to address the identity learning inefficiencies found in the existing identity loss.
3) A training-free method to adapt the model to multi-reference inputs during inference.
The proposed method demonstrably restores high-quality faces and achieves state-of-the-art identity preserving restoration on benchmarks such as FFHQ-Ref and CelebA-Ref-Test, consistently outperforming previous work.




\end{abstract}

\section{Introduction}

%
Recently, image restoration~\cite{esrgan,stablesr,supir,diffbir,seesr,pasd} has seen significant improvements
along with the rise of diffusion models~\cite{ddpm,ddim}, particularly in terms of
generated image quality~\cite{ldm,sdxl}.
However, the state-of-the-art restoration methods, including the face-specific
ones~\cite{codeformer,diffbir,refldm,restoreid}, still suffer from
unsatisfactory identity preservation when processing facial imagery.
This limitation can substantially degrade the user experience, given the human perceptual acuity for subtle variations in facial features.
%

%
In some real-world applications, such as digital albums, when restoring a
low-quality face image, it is possible to leverage other
high-quality face images from the same person
as references to better preserve the identity and appearance.
Consequently, many reference-based face restoration methods~\cite{refldm,restoreid,instantrestore,dmdnet} have been proposed.
These efforts involve
designing novel architectures for reference face conditioning~\cite{restoreid,refldm},
formulating new loss functions for identity preservation and
image quality~\cite{refldm,instantrestore}, and curating specialized reference-based face restoration datasets~\cite{refldm,dmdnet}.
Nevertheless, the existing methods do not fully exploit the potential
of reference faces, and hence there is room for improvement in both identity
preservation and image quality.

In this paper, to more effectively utilize the reference face images
and further enhance the performance of reference-based
face restoration,
we propose two independent modules that exploit the reference face
in two different aspects: representation and supervision.

First, we propose Composite Context, a comprehensive representation
for the reference face.
It consists of multiple pre-trained face representations that
focus on different information in the reference face from high-level to low-level.
Specifically, it includes identity embedding~\cite{arcface} as high-level identity information; and general face representation~\cite{farl} comprising both high-level semantic information and low-level face information.
%
These comprehensive and multi-level representations can
exploit the reference face images better than
previous work that only
uses a single type of representation~\cite{restoreid,osdface,refldm}, capturing only partial information which is insufficient.
For example, the face embedding~\cite{facenet,arcface}
only contains identity information, while other potentially
useful details like the visual appearance of the reference face
are not encoded in face embedding at all.
%
%
%

Second, we propose Hard Example Identity Loss.
%
%
It is a simple yet effective extension of the existing identity loss~\cite{refldm}, motivated by the empirical observation that traditional identity loss functions suffer from learning inefficiencies -- a well-documented issue in the metric learning literature~\cite{facenet,dmlcheck,dmlrevisit}.
In particular, the ground-truth faces are not hard enough (see ``Triplet Selection'' in~\cite{facenet} for the meaning of ``hard example''), which makes the identity loss magnitude
very small after a short period of training.
By simply incorporating a hard sample, namely the reference face,
into the identity loss, the learning inefficiency problem can be
effectively addressed, and hence leads to a significant 
performance
improvement.
%
In contrast, all previous works~\cite{osdface,refldm,instantrestore} overlooked this issue.
%
%
%

Apart from the representation and supervision aspects, while our  method is designed to take a single
reference face image, it can be extended to support multiple reference face images
through a simple method based on classifier-free guidance~\cite{cfg}
at the inference stage, which requires no extra training.

Our qualitative and quantitative experimental results on the FFHQ-Ref~\cite{refldm}
and CelebA-Ref-Test~\cite{refldm} datasets
demonstrate the effectiveness
of our method.
Though simple, our method effectively and consistently outperforms the previous methods in face identity preservation.

\textbf{Contributions}.
Our contributions are threefold regarding reference-based face image restoration:
\begin{itemize}[nosep,leftmargin=*]
\item We propose ``Composite Context'', a comprehensive face representation that involves multi-level information extracted from the reference face.

\item We propose ``Hard Example Identity Loss'', an extension to the existing identity loss
that incorporates the reference face for improved learning efficiency and identity
preservation.
\item We extend the proposed model
to flexibly support multi-reference faces at the inference stage
in a training-free manner, even though the model only uses a single reference face
during training.
\end{itemize}
%
%

\section{Related Work}

\noindent
\textbf{Image Restoration.}
As diffusion models~\cite{ddpm,ldm,ddim,consistencymodel,adm,sde} gain popularity in image generation,
LDM~\cite{ldm} has recently become a popular backbone for general image
restoration~\cite{stablesr,diffbir,supir,pasd,seesr,mmsr}.
However,  humans are perceptually highly sensitive to subtle differences in face images, general image restoration techniques typically perform poorly, especially in terms of identity preservation and maintaining face
image realism.
%
%
In this case, face-specific restoration models are preferred.

\noindent
\textbf{No-reference Face Restoration.}
When there is no reference face, generative models can be used to hallucinate
details while restoring a degraded facial image~\cite{codeformer,diffbir,osdface,fsrnet,asffnet,gfpgan,pgdiff,dfdnet}.
CodeFormer~\cite{codeformer} presents a Transformer~\cite{transformer} to model the global composition
and context of the low-quality faces for code prediction, enabling the generation of
natural faces that closely approximate the target faces.
DiffBIR~\cite{diffbir} presents a two-stage pipeline for blind face restoration,
involving the degradation removal stage and information regeneration stage.
OSDFace~\cite{osdface} proposes a visual representation embedder to capture
information from the low-quality input face and incorporate the face identity loss
for identity preservation.
A common challenge in no-reference face restoration is identity preservation
since no additional information is provided.

\noindent
\textbf{Reference-based Face Restoration}.
High-quality reference face images, when available,
can help identity preservation when restoring a low-quality face of
the same person~\cite{pdgrad,refldm,restoreid,instantrestore,dmdnet,pfstorer}.
%
DMDNet~\cite{dmdnet} proposes a dual memory dictionary for both general and identity-specific
features for blind face restoration.
RestorerID~\cite{restoreid} presents a Face ID Adapter and incorporates the identity embedding
of the reference face as a tuning-free face restoration method.
InstantRestore~\cite{instantrestore} leverages a one-step diffusion model, and proposes
a landmark attention loss to enhance identity preservation.
RefLDM~\cite{refldm} incorporates the CacheKV mechanism and
a timestep-scaled identity loss into an LDM~\cite{ldm} to effectively utilize multiple reference
faces for face restoration.
Personalization methods~\cite{pfstorer,faceme}
utilizes reference faces with the goal of customizing the model for individual users.

\textbf{Key Differences.}
This paper focuses on reference-based face restoration.
The most related works to this paper are RefLDM~\citep{refldm}, RestorerID~\cite{restoreid},
and InstantRestore~\cite{instantrestore}.
The key differences between our method and the previous methods are:
\textbf{(1)}
The previous methods only use a single representation for the reference face,
which only covers partial information of the reference and does not
maximize the utilization of the reference.
%
%
%
In contrast, our Composite Context combines multi-level face-specific representations
to comprehensively exploit the information in the reference face.
The Composite Context conceptually resembles \cite{mmsr,llvfsr} which employ multiple modalities to aid image restoration,
and \cite{sdxl} which concatenates two text representations for text-to-image synthesis.
\textbf{(2)}
While many related works~\cite{refldm,osdface,instantrestore} incorporate the identity loss,
a notable learning inefficiency issue where the loss value plateaus at a tiny value (indicates learning inefficiency
in the context of metric learning~\cite{dmlcheck,dmlrevisit,facenet}) has been overlooked.
We propose Hard Example Identity Loss to incorporate ``hard examples'' to address this issue.
%
\textbf{(3)}
Some existing methods support only one reference face~\cite{restoreid}.
Some others support multiple~\cite{refldm}, but also
require multiple reference face images during training.
It is noted that requiring more than one reference face makes
training data collection difficult.
In contrast, even if our method only uses one reference face
during training, our model can be adapted to support
multiple reference faces during inference in a training-free manner.

\section{Our Approach}

Given a low-quality (LQ) face image $\bm{x}_\text{LQ}$, and a 
high-quality (HQ) reference face image $\bm{x}_\text{REF}$ from the same
person, we aim to restore the LQ image while preserving the person identity
by leveraging the reference face.
The resulting image should be close to the ground truth HQ image $\bm{x}_\text{HQ}$
in terms of both identity similarity and perceptual similarity.

To this end, we adopt a general LDM~\cite{ldm} backbone pretrained
for text-to-image synthesis.
Following the previous works~\cite{ldm,refldm}, we incorporate the
LQ input image $\bm{x}_\text{LQ}$ by conditioning the diffusion model
on its corresponding VAE latent $\bm{z}_\text{LQ}$ through
concatenating it to the noise latent $\bm{z}_t$.
In this way, the model $\bm{\epsilon}(\bm{z}_t,\bm{z}_\text{LQ},t)$
can serve as a fundamental face image restoration model.

\begin{figure}
    \centering
    \includegraphics[width=\linewidth]{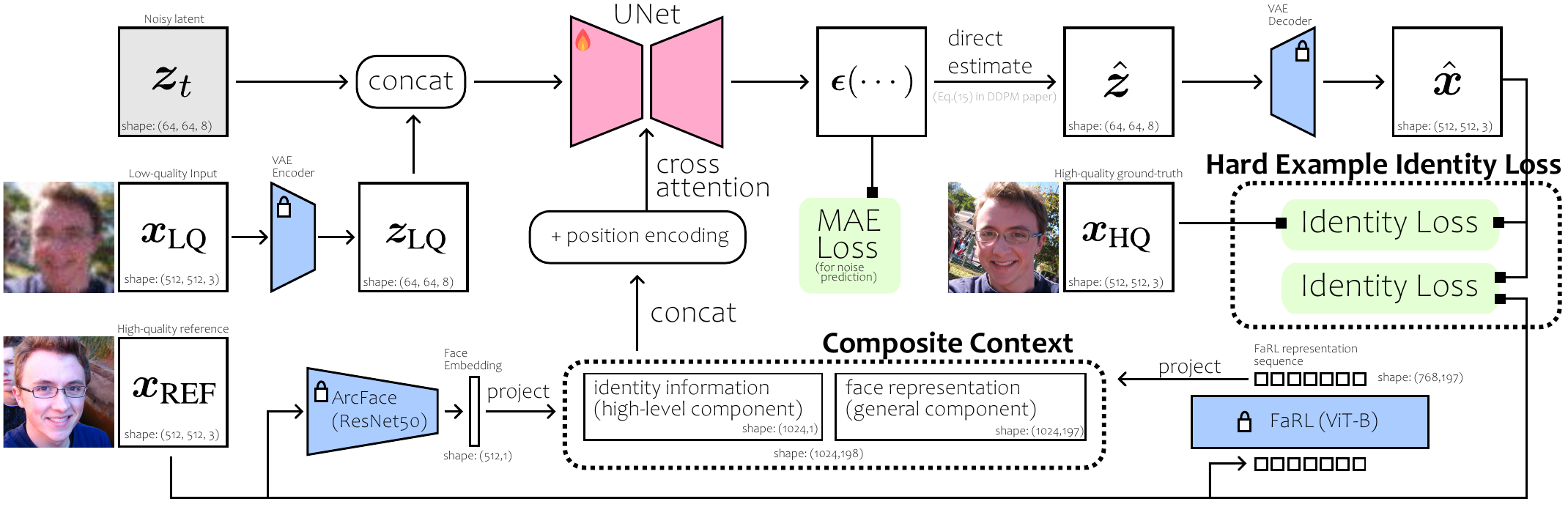}
    \vspace{-2em}
    \caption{Overview of our proposed method. The Composite Context and Hard Example Identity Loss are designed for fully exploiting the reference face and hence boost identity preservation.
    The $\bm{z}_t$ is the noisy latent, $\bm{x}_\text{LQ}$ is
    the low-quality face image input ($\bm{z}_\text{LQ}$ is its
    corresponding VAE latent), $\bm{x}_\text{REF}$ is
    the high-quality reference face, $\bm{x}_\text{HQ}$ is
    the high-quality ground-truth face image,
    $\hat{\bm{z}}$ is the direct estimate of the denoised
    result (\emph{i.e.}, Eq.~(15) in DDPM~\cite{ddpm}),
    and $\hat{\bm{x}}$ is the VAE decoded direct estimate.
    All pre-trained modules are frozen. The UNet~\cite{unet}
    and projection matrices for Composite Context are trained.
    The total loss includes the MAE loss and the
    Hard Example Identity Loss.
    }
    \label{fig:overview}
\end{figure}

In order to comprehensively leverage the reference face for better
identity preservation, we propose two independent modules:
Composite Context (CC) and Hard Example Identity Loss (HID), which will be detailed
in Section.~\ref{sec:31} and Section.~\ref{sec:32} below.
In brief, the Composite Context is a comprehensive
representation $\bm{c}$ from the reference face $\bm{x}_\text{REF}$.
It is used as a condition for $\bm{\epsilon}(\bm{z}_t,\bm{z}_\text{LQ},\bm{c},t)$ through cross-attention mechanism~\cite{ldm}.
Then, the Hard Identity Loss $\mathcal{L}_\text{HID}$ will
take advantage of the reference face to enhance identity
preservation.
See Figure~\ref{fig:overview} for the overview of our method.

\subsection{Composite Context for Comprehensive Reference Face Representation}
\label{sec:31}

Different from no-reference face restoration methods,
reference-based face restoration~\cite{refldm,restoreid,instantrestore}
assumes that a high-quality
reference face from the same person is available.
To thoroughly leverage this advantage, we propose
Composite Context, a comprehensive representation of the
reference face image that covers multi-level information
from the reference face, including high-level semantic
information (such as person identity) and low-level
appearance information (such as skin texture).
Unlike previous works~\cite{osdface,refldm,restoreid,instantrestore} that only leverage partial information
from the reference face through a single representation,
Composite Context allows the model to comprehensively leverage
the reference face at different levels.
Therefore, Composite Context may benefit identity preservation.

Given a reference face image $\bm{x}_\text{REF}$ which belongs to
the same identity as $\bm{x}_\text{LQ}$, we can leverage a collection
of pre-trained face representation models for various purposes to
extract the respective representations, and combine them together as
a vector sequence.
In particular, Composite Context consists of the following multi-level
components:
\begin{itemize}[itemsep=0pt,leftmargin=*]

    \item \textbf{High-level features:} ArcFace~\cite{arcface} embedding representing
    person identity.
    ArcFace is a face recognition model which enforces an angular margin
    loss in its embedding space.
    We assume that $\phi_\text{H}(\cdot)$
    is the pre-trained ArcFace model in the standard ResNet50~\cite{resnet} architecture, and $\bm{W}_\text{H}$ is a
    projection matrix from the dimensionality of face embedding
    to the dimension of UNet cross-attention.
    The projected embedding
    $\bm{W}_\text{H}\phi_\text{H}(\bm{x}_\text{REF})$
    is the first part of the Composite Context.
    
    \item \textbf{General features:} FaRL~\cite{farl} representation representing
    various high-level semantic (\textit{e.g.}, face attributes) and low-level information (\textit{e.g.}, visual appearance) of the reference face.
    FaRL is a general face representation model learned in a
    visual-linguistic manner, with image-text contrastive
    learning and masked image modeling simultaneously~\cite{farl}.
    We assume that $\phi_\text{G}(\cdot)$ is a pre-trained FaRL model (ViT-B~\cite{vit}
    architecture), and $\bm{W}_\text{G}$ is the projection matrix
    to the dimention of UNet cross-attention.
    We use the whole output sequence (197 tokens) from FaRL
    to maximize reference face utility.
    The projected sequence $\bm{W}_\text{G}\phi_\text{G}(\bm{x}_\text{REF})$
    is the second part of the Composite Context.
    %
    
    %
    %
    %
    
\end{itemize}
After obtaining those different representations from the reference
face $\bm{x}_\text{REF}$, they are concatenated, and added with the standard sinusoidal
positional encoding~\cite{transformer} as the Composite Context:
\begin{equation}
    \bm{c} = \text{Concat}\big[
    \bm{W}_\text{H}\phi_\text{H}(\bm{x}_\text{REF}),
    \bm{W}_\text{G}\phi_\text{G}(\bm{x}_\text{REF})
    \big] + \bm{e}_\text{position},
\end{equation}
where $\bm{e}_\text{position}$ denotes sinusoidal positional encoding~\cite{transformer}.
Since all the Composite Context components are from pre-trained models,
the sequence length is fixed at $1+197=198$ for any reference face.
Finally, the Composite Context $\bm{c}$ is incorporated into the model
through the cross-attention conditioning mechanism~\cite{ldm} as
$\bm{\varepsilon}(\bm{z}_t,\bm{z}_\text{LQ},\bm{c},t)$.
See Figure~\ref{fig:overview} for the overall diagram.


\subsection{Hard Example Identity Loss for Improved Learning Efficiency}
\label{sec:32}


One of the goals of face restoration is to preserve the identity,
which means the restored face should match the identity of the HQ image.
To achieve this,
many recent works~\cite{refldm,osdface,instantrestore} incorporate the identity
loss, which is based on a pre-trained face embedding model~\cite{arcface,facenet} such as ArcFace~\cite{arcface}.
In particular, RefLDM~\cite{refldm} presents a timestep-scaled identity loss
$\mathcal{L}_\text{ID}$ as:
\begin{equation}
\mathcal{L}_\text{ID}(\bm{x}_\text{HQ}, \hat{\bm{x}}) =
\sqrt{\bar{\alpha}_t} \cdot \big( 1 - \cos \langle\phi_\text{H}(\bm{x}_\text{HQ}),
\phi_\text{H}(\hat{\bm{x}}) \rangle
\big),
\label{eq:idloss}
\end{equation}
where $\phi_\text{H}$ denotes the face embedding model~\cite{arcface},
the notation $\sqrt{\bar{\alpha}_t}$ is inherited from DDPM~\cite{ddpm},
and $\hat{\bm{x}}$ is the direct estimate of $\bm{x}_0$ at time step $t$, \textit{i.e.}, Eq.~(15) in DDPM~\cite{ddpm}.
The time-step scaling factor $\sqrt{\bar{\alpha}_t}$ mitigates
the out-of-domain behavior of the identity loss at a very noisy step $t$,
and emphasizes identity preservation at less noisy steps.
%
%
However, a learning inefficiency issue is overlooked.

\begin{wrapfigure}{r}{0.35\linewidth}
\vspace{-1.5em}
\includegraphics[width=\linewidth]{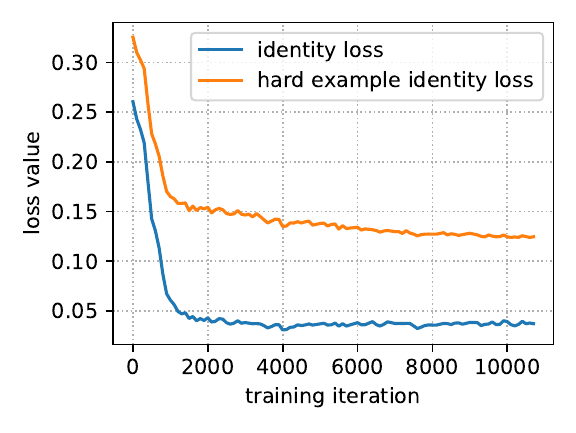}
\vspace{-2em}
\caption{Loss curves of $\mathcal{L}_\text{ID}$ and $\mathcal{L}_\text{HID}$
during the training process.
The curves are truncated to the beginning part of the training process.
}
\vspace{-0.7em}
\label{fig:losscurve}
\end{wrapfigure}

During experiments, we observe that the identity loss in Eq.~\eqref{eq:idloss}
decreases quickly and plateaus at a very small magnitude, as shown by the
blue curve in Figure~\ref{fig:losscurve}.
In the metric learning literature~\cite{facenet,dmlcheck,dmlrevisit,robdml,robrank},
there is a similar phenomenon where the loss value is small
when the training samples are not hard enough (see ``Triplet Selection'' in~\cite{facenet}), which usually leads to poor
generalization.
Their countermeasure is to mine some hard examples~\cite{facenet} that
can trigger a larger loss value so the model performance can be drastically
influenced~\cite{dmlrevisit}.
Inspired by such solution to the learning inefficiency issue, we propose to leverage the reference face $\bm{x}_\text{REF}$
as a hard example in addition to $\bm{x}_\text{HQ}$.
Based on this, we design a simple extension to the  identity loss
$\mathcal{L}_\text{ID}$
as the ``Hard Example Identity Loss'' incorporating the hard example, namely the reference face.

Let $\lambda$ be a hyper-parameter for balancing the influence of $\bm{x}_\text{HQ}$
and $\bm{x}_\text{REF}$ during training.
Formally, the Hard Example Identity Loss $\mathcal{L}_\text{HID}$  is also based
on the direct estimate $\hat{\bm{x}}$, and is defined as:
\begin{equation}
\mathcal{L}_\text{HID}(\bm{x}_\text{HQ}, \bm{x}_\text{REF}, \hat{\bm{x}}) =
(1-\lambda) \mathcal{L}_\text{ID}(\bm{x}_\text{HQ}, \hat{\bm{x}})
+ \lambda \mathcal{L}_\text{ID}(\bm{x}_\text{REF}, \hat{\bm{x}}).
\label{eq:hid}
\end{equation}
As shown by the orange curve in Figure~\ref{fig:losscurve},
our Hard Example Identity Loss will no longer plateau at a very small value
because a ``harder'' example is introduced, and hence
will alleviate the learning inefficiency issue.
While simple in its form, the introduction of the reference face is very
effective and can clearly improve the identity preservation.
%
%
As a different interpretation of the introduction of the reference face, 
it is noted that the input faces are noisy
(as they are direct estimations during DDPM),
which inherently makes the face embedding and the identity loss noisy.
%
In this case, introducing the additional contrastiveness through the reference face
can potentially lead to a regularization effect, stabilizing the gradients from the identity loss.
%
%
The total loss of our model is the L-1 diffusion loss (\emph{aka.} MAE)
and the Hard Example Identity Loss with a balancing hyper-parameter $w_\text{HID}$:
\begin{equation}
    \mathcal{L}_\text{total} = \mathcal{L}_\text{MAE}
    + w_\text{HID}\cdot \mathcal{L}_\text{HID}.
\end{equation}

\begin{figure}[t]

\noindent%
\setlength{\tabcolsep}{0.2mm}%
\renewcommand{\arraystretch}{0.4}%
\resizebox{\linewidth}{!}{%
\begin{tabular}{cccccccc}

\cellcolor{gray!20} LQ & \cellcolor{gray!20} REF & CodeFormer~\cite{codeformer} & DiffBIR~\cite{diffbir} & RefLDM~\cite{refldm} & RestorerID~\cite{restoreid} & \cellcolor{yellow!50} \textbf{Ours} & \cellcolor{gray!20} HQ \\

\midrule

\insertimagescontextirmoderatevtwo{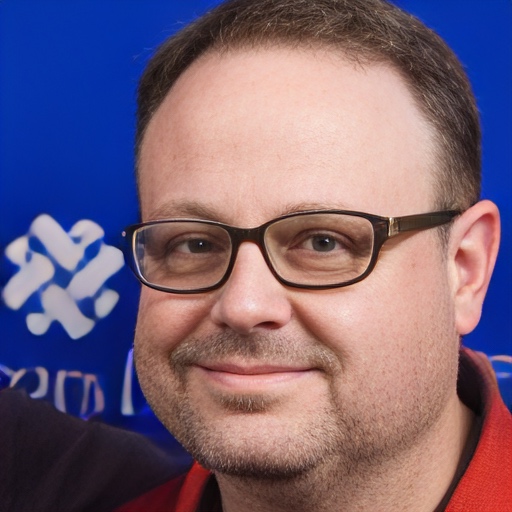} \\

\insertimagescontextirmoderatevtwo{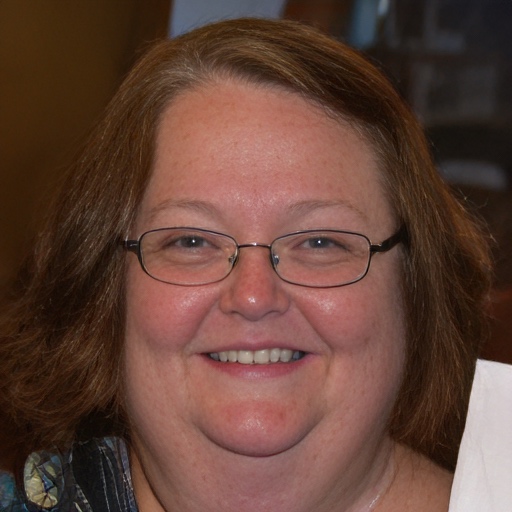} \\

\insertimagescontextirmoderatevtwo{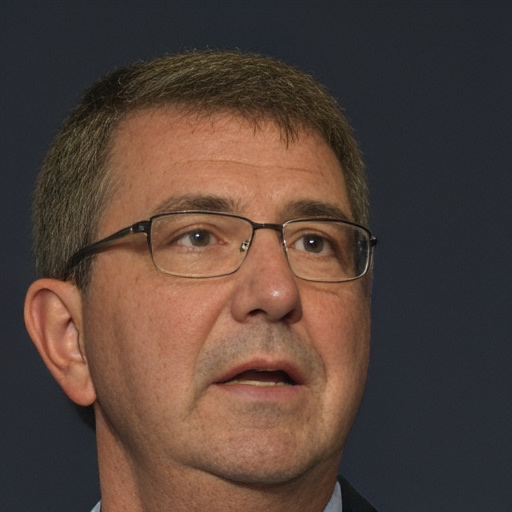} \\

\insertimagescontextirmoderatevtwo{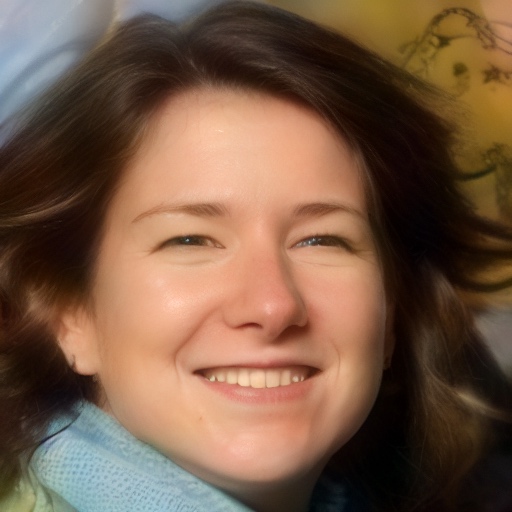} \\

\insertimagescontextirmoderatevtwo{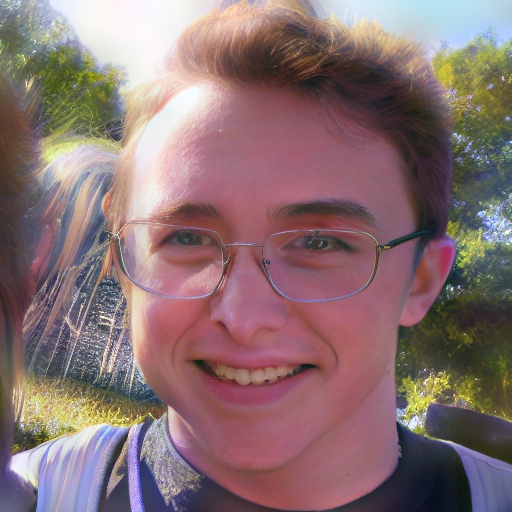} \\

\insertimagescontextirmoderatevtwo{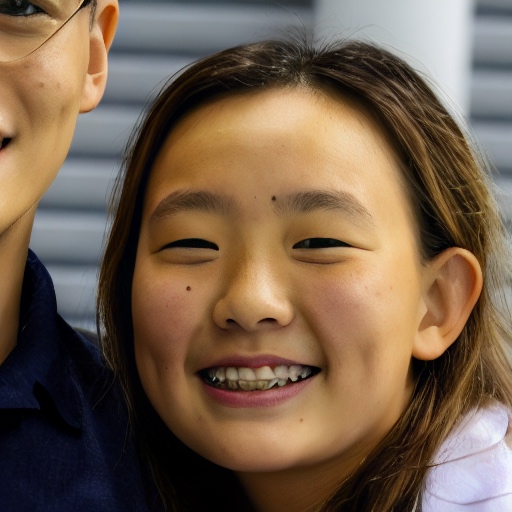} \\

\midrule

\cellcolor{gray!20} LQ & \cellcolor{gray!20} REF & CodeFormer~\cite{codeformer} & DiffBIR~\cite{diffbir} & RefLDM~\cite{refldm} & RestorerID~\cite{restoreid} & \cellcolor{yellow!50} \textbf{Ours} & \cellcolor{gray!20} HQ \\

\end{tabular}}

\vspace{-0.3em}
\caption{Qualitative comparison with other state-of-the-art face restoration methods on FFHQ-Ref Moderate~\cite{refldm} test set. The ``REF''
column is the reference face.
Please zoom in for the face details.
For instance, the black moles are well-preserved in our result on the sixth row.
}
\label{fig:fig1}
\end{figure}

\subsection{Training-Free Extension for Multi-Reference Faces}


%
Classifier-free guidance~\cite{cfg} is an effective technique for improving diffusion model
performance, which is also widely adopted in the image restoration literature~\cite{diffbir,supir,stablesr}.
Since our model involves both the LQ condition $\bm{z}_\text{LQ}$ and $\bm{c}$, we follow \cite{ip2p} for
their classifier-free guidance formulation:
\begin{equation}
\tilde{\bm{\epsilon}}(\bm{z}_t,\bm{z}_\text{LQ}, \bm{c}, t) =
(1 - s_i) \bm{\epsilon}(\bm{z}_t, \varnothing, \varnothing, t)
+ (s_i - s_c) \bm{\epsilon}(\bm{z}_t, \bm{z}_\text{LQ}, \varnothing, t)
+ s_c \bm{\epsilon}(\bm{z}_t,\bm{z}_\text{LQ}, \bm{c}, t),
\label{eq:cfg}
\end{equation}
where $s_c$ controls the guidance effect of the composite context $\bm{c}$,
and $s_i$ controls the guidance effect of the LQ latent $\bm{z}_\text{LQ}$.
The two hyper-parameters $s_i$ and $s_c$ can be adjusted
at the inference stage.
%
%

%
While our method is designed to take only one reference face image,
it can be extended to support multiple reference faces through
a simple ensemble.
Let $\bm{C}=\{\bm{c}_i\}_{i=1,\ldots,N}$ be a set of composite contexts obtained from $N$ reference face images.
The multi-reference inference is formulated as:
\begin{equation}
\tilde{\bm{\epsilon}}(\bm{z}_t,\bm{z}_\text{LQ}, \bm{C}, t) =
(1 - s_i) \bm{\epsilon}(\bm{z}_t, \varnothing, \varnothing, t)
+ (s_i - s_c) \bm{\epsilon}(\bm{z}_t, \bm{z}_\text{LQ}, \varnothing, t)
+ \frac{s_c}{N} \sum_{i=1}^N
\bm{\epsilon}(\bm{z}_t,\bm{z}_\text{LQ}, \bm{c}_i, t).
\label{eq:multicfg}
\end{equation}
Inspired by \cite{refldm}, the identity is expected to be
better preserved when more reference faces are provided.
Different from \cite{refldm} which uses multiple reference
faces for training, our method only requires
one reference face during training while being able to use
multiple reference faces during inference.
Our method alleviates the data scarcity
issue in the multi-reference face scenario, where most training
samples only have a single reference face~\cite{refldm}.
Such paradigm could be scalable.

%
%
%
%

\begin{figure}[t]

\noindent%
\setlength{\tabcolsep}{0.2mm}%
\renewcommand{\arraystretch}{0.4}%
\resizebox{\linewidth}{!}{%
\begin{tabular}{cccccccc}

\cellcolor{gray!20} LQ & \cellcolor{gray!20} REF & CodeFormer~\cite{codeformer} & DiffBIR~\cite{diffbir} & RefLDM~\cite{refldm} & RestorerID~\cite{restoreid} & \cellcolor{yellow!50} \textbf{Ours} & \cellcolor{gray!20} HQ \\

\midrule

\insertimagescontextirseverevtwo{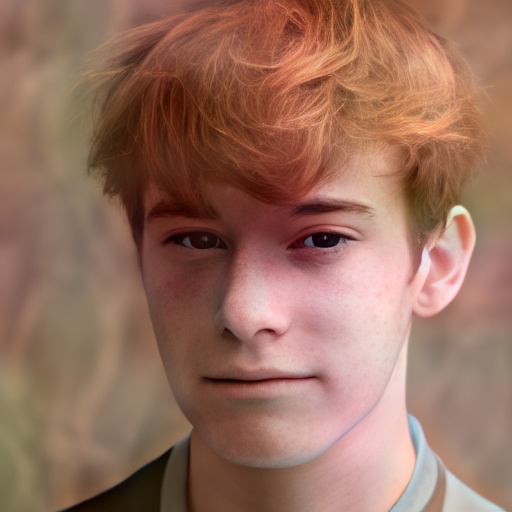} \\

\insertimagescontextirseverevtwo{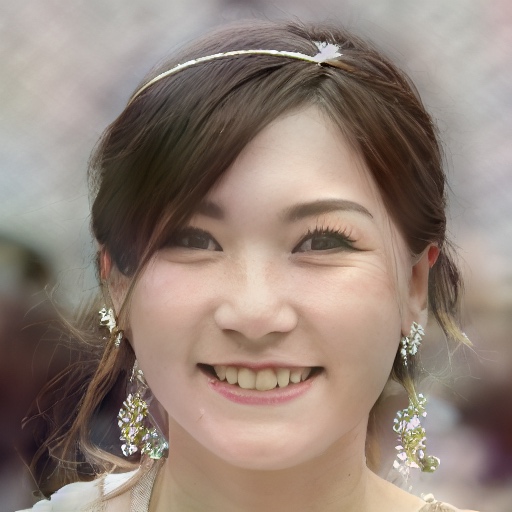} \\

\insertimagescontextirseverevtwo{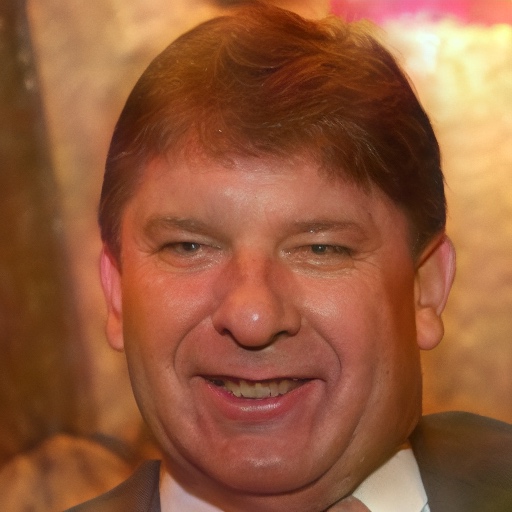} \\

\insertimagescontextirseverevtwo{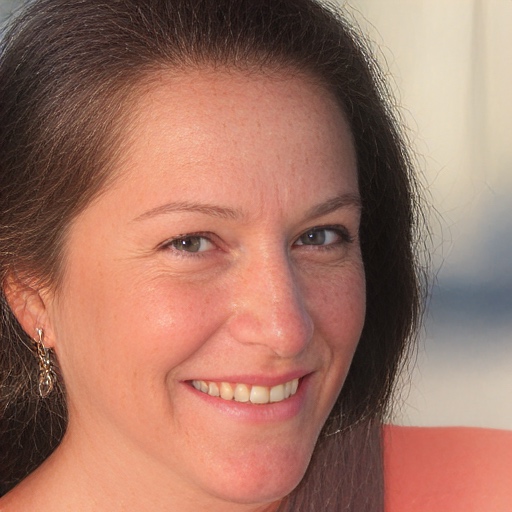} \\

\insertimagescontextirseverevtwo{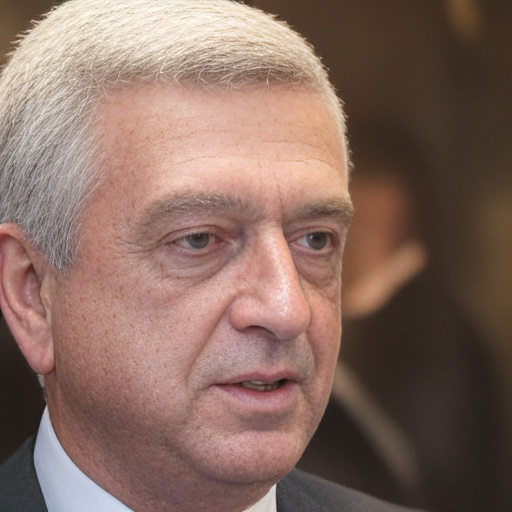} \\

\insertimagescontextirseverevtwo{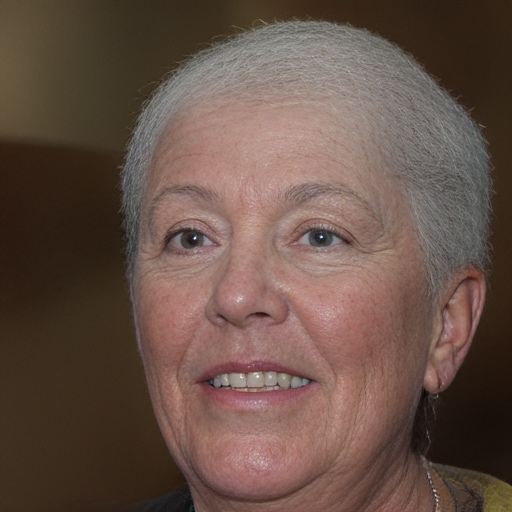} \\

\midrule

\cellcolor{gray!20} LQ & \cellcolor{gray!20} REF & CodeFormer~\cite{codeformer} & DiffBIR~\cite{diffbir} & RefLDM~\cite{refldm} & RestorerID~\cite{restoreid} & \cellcolor{yellow!50} \textbf{Ours} & \cellcolor{gray!20} HQ \\

\end{tabular}}

\vspace{-0.3em}
\caption{Qualitative comparison with other state-of-the-art face restoration methods on FFHQ-Ref Severe~\cite{refldm} test set. The ``REF'' column is the high-quality reference face image.}
\label{fig:fig2}
\end{figure}

\section{Experiments}

\noindent\textbf{Datasets.}
Our model is trained on the FFHQ-Ref~\cite{refldm} dataset,
which is a subset of FFHQ~\cite{stylegan} by person identity
clustering.
It comprises $18816$ images for training and $857$ images for testing.
We follow \cite{realesrgan} 
for their second-order degradation simulation pipeline during training.
For training data augmentation, we use random horizontal flipping with $0.5$ probability,
and random color jittering with $0.5$ probability.
For testing purposes, we adopt the identical test data from \cite{refldm}, namely
FFHQ-Ref Moderate, FFHQ-Ref Severe, and CelebA-Ref-Test~\cite{refldm}.
In this paper, the face image resolution is always $512\times 512$ following previous
works~\cite{codeformer,diffbir,refldm,restoreid}.
Note, while most previous works do not use identical training data and may
potentially suffer from
test data leakage~\cite{refldm}, our training and test
images are completely identical to RefLDM~\cite{refldm} (NeurIPS'24)
for a fair comparison.

\noindent\textbf{Implementation Details.}
We employ an LDM~\cite{ldm} backbone with 865M parameters pre-trained
on the WebLI~\cite{pali} dataset for text-to-image synthesis.
We fine-tuned the VAE following~\cite{refldm}, using the $68411$ remaining FFHQ~\cite{stylegan} images after excluding the FFHQ-Ref~\cite{refldm}
validation and test images.
Our model is trained on the FFHQ-Ref training set for 100K steps,
with batch size $256$ and learning rate \verb|8e-5|.
The cross-attention dimension is $1024$.
To enable classifier-free guidance~\cite{cfg,ip2p}, we randomly drop
the LQ condition as well as the components in Composite Context independently
with a $0.1$ probability. The Composite Context components are dropped
through attention masking.
The classifier-guidance scales are selected as $s_i=1.2$ and $s_c=1.2$  for inference. 
The Hard Example Identity Loss balancing parameter $w_\text{HID}$ is $0.1$ following
\cite{refldm}, and $\lambda$ is set as $0.6$ by default.
The AdaIN~\cite{stylegan}-based color fix~\cite{stablesr} is applied
on the model output as a post-processing step.
%
%

\noindent\textbf{Evaluation.}
Following the previous works~\cite{refldm,restoreid,instantrestore},
we use LPIPS~\cite{lpips} for perceptual similarity,
and IDS (\textit{i.e.}, the cosine similarity of ArcFace~\cite{arcface} embedding) for
person identity preservation.
This ``IDS'' is calculated between the restoration result
and the HQ image.
Since we optimize the identity loss using the ArcFace~\cite{arcface} model during training, using IDS
alone may not properly reflect generalization performance due to potential
overfitting.
Thus, we also evaluate the ArcFace IDS with respect to the first reference
face for each LQ test image (denoted as ``IDS(REF)''), as well as the FaceNet
IDS with respect to HQ (denoted as ``FaceNet'').
We also use no-reference metrics including MUSIQ~\cite{musiq}, NIQE~\cite{niqe},
and FID~\cite{fid}
for image quality evaluation.

\subsection{Experimental Results and Comparison with SOTA}


\begin{table}[t]

\caption{Comparison with state-of-the-art face restoration methods on FFHQ-Ref
Moderate and Severe~\cite{refldm}.
The ``\#REF'' means the number of reference face used.}
\label{tab:tab1}

\vspace{-0.5em}
\resizebox{\linewidth}{!}{
\setlength{\tabcolsep}{2pt}%
\begin{tabular}{cc|ccccccc|ccccccc}
\toprule 
\multirow{2}{*}{\textbf{Method}} & \multirow{2}{*}{\textbf{\#REF}} & \multicolumn{7}{c|}{\textbf{FFHQ-Ref Moderate}} & \multicolumn{7}{c}{\textbf{FFHQ-Ref Severe}}\tabularnewline
\cmidrule{3-16}
 &  & IDS$\uparrow$ & FaceNet$\uparrow$ & IDS(REF)$\uparrow$ & LPIPS$\downarrow$ & MUSIQ$\uparrow$ & NIQE$\downarrow$ & FID$\downarrow$ & IDS$\uparrow$ & FaceNet$\uparrow$ & IDS(REF)$\uparrow$ & LPIPS$\downarrow$ & MUSIQ$\uparrow$ & NIQE$\downarrow$ & FID$\downarrow$\tabularnewline
\midrule 
CodeFormer (NeurIPS'22) & 0 & 0.783 & 0.822 & 0.545 & \textbf{0.1839} & \uline{75.88} & \uline{4.38} & 31.7 & 0.370 & 0.677 & 0.265 & \textbf{0.3113} & \textbf{76.12} & \uline{4.30} & 49.6\tabularnewline
DiffBIR (ECCV'24) & 0 & \uline{0.831} & \uline{0.842} & 0.575 & 0.2268 & \textbf{76.64} & 5.72 & 34.9 & 0.356 & 0.672 & 0.253 & 0.3606 & \textbf{75.71} & 6.24 & 55.3\tabularnewline
\midrule 
RefLDM (NeurIPS'24) & 1 & 0.826 & 0.837 & \uline{0.624} & 0.2211 & 72.30 & 4.61 & \uline{28.0} & \uline{0.571} & \uline{0.733} & \uline{0.554} & \uline{0.3366} & 74.32 & 4.52 & \textbf{36.0}\tabularnewline
RestorerID (arXiv) & 1 & 0.804 & 0.832 & 0.591 & 0.2350 & 73.35 & 4.98 & 31.0 & 0.411 & 0.690 & 0.408 & 0.4130 & 74.49 & 4.71 & 52.7\tabularnewline
\midrule 

Ours & 1 & \textbf{0.843} & \textbf{0.850} & \textbf{0.732} & \uline{0.2054} & 75.29 & \textbf{3.96} & \textbf{25.5}
& \textbf{0.609} & \textbf{0.743} & \textbf{0.712} & 0.3647 & 75.22 & \textbf{3.84} & \uline{38.3}\tabularnewline

\bottomrule
\end{tabular}}
\vspace{-0.5em}

\end{table}

To validate the effectiveness of our proposed method, we 
evaluate our method on the FFHQ-Ref test datasets with Moderate and Severe degradations, and CelebA-Ref-Test following~\cite{refldm}.
We compare our method with some state-of-the-art
no-reference face restoration methods, namely CodeFormer~\cite{codeformer}
and
DiffBIR~\cite{diffbir}, as well as the latest reference-based face
restoration methods, namely
RefLDM~\cite{refldm} and RestorerID~\cite{restoreid}.
The quantitative results on FFHQ-Ref test datasets can be found in Table~\ref{tab:tab1}.
The multi-reference results are in Table~\ref{tab:multiref}.
The quantitative results on CelebA-Ref-Test can be found in Table~\ref{tab:tab1-celeb}.
The visualization for FFHQ-Ref test sets can be found in
Figure~\ref{fig:fig1} and Figure~\ref{fig:fig2}.
All results of the related works are reproduced using their
official code and checkpoints.
At the time of writing, some other related works such as
OSDFace~\cite{osdface} and InstantRestore~\cite{instantrestore}
have not yet published their code and checkpoints.
Hence they are not included for comparison.

As shown in Table~\ref{tab:tab1}, 
the IDS and FaceNet are computed between the output and HQ ground-truth,
whereas IDS(REF) is computed between the output
and the first reference face.
The overall trend is that no-reference methods like CodeFormer~\cite{codeformer} and DiffBIR~\cite{diffbir}
tend to achieve good perceptual similarity (LPIPS) and
image quality (MUSIQ),
but worse identity preservation compared to reference-based methods like
RefLDM~\cite{refldm} and RestorerID~\cite{restoreid}.
And notably, our model consistently achieves the best identity preservation
(which is the top-priority in the reference-based face restoration task)
across all test datasets,
while still achieving competitive image quality.

\begin{table}[t]

\caption{Multi-reference face inference result on FFHQ-Ref test sets. The identity preservation improves when the number of reference faces increases~\cite{refldm}. Note, the IDS(REF) is calculated using the first reference face, and it may drop with more than one reference face, because the additional reference faces can pull the model output slightly further from the first reference face in Eq.~\eqref{eq:multicfg}.}
\label{tab:multiref}

\vspace{-0.5em}
\resizebox{\linewidth}{!}{
\setlength{\tabcolsep}{2pt}%
\begin{tabular}{c|ccccccc|ccccccc}
\toprule 
\multirow{2}{*}{\textbf{\#REF}} & \multicolumn{7}{|c|}{\textbf{FFHQ-Ref Moderate}} & \multicolumn{7}{c}{\textbf{FFHQ-Ref Severe}}\tabularnewline
\cmidrule{2-15}
 & IDS$\uparrow$ & FaceNet$\uparrow$ & IDS(REF)$\uparrow$ & LPIPS$\downarrow$ & MUSIQ$\uparrow$ & NIQE$\downarrow$ & FID$\downarrow$ & IDS$\uparrow$ & FaceNet$\uparrow$ & IDS(REF)$\uparrow$ & LPIPS$\downarrow$ & MUSIQ$\uparrow$ & NIQE$\downarrow$ & FID$\downarrow$\tabularnewline
\midrule

1 & 0.843 & 0.850 & \textbf{0.732} & 0.2054 & \textbf{75.29} & {3.96} & 25.5
  & 0.609 & 0.743 & \textbf{0.712} & 0.3647 & \textbf{75.22} & 3.84 & 38.3
\tabularnewline

2 & 0.857 & 0.856 & 0.693 & 0.2042 & 75.28 & \textbf{3.95} & \textbf{25.4}
  & 0.640 & 0.752 & 0.650 & 0.3625 & 75.20 & \textbf{3.82} & \textbf{38.2}
  \tabularnewline

3 & 0.861 & \textbf{0.859} & 0.683 & 0.2040 & \textbf{75.29} & {3.96} & \textbf{25.4}
  & 0.652 & 0.755 & 0.636 & 0.3619 & 75.19 & \textbf{3.82} & 38.4
  \tabularnewline

4 & \textbf{0.863} & \textbf{0.859} & 0.680 & 0.2039 & \textbf{75.29} & {3.96} & 25.5
  & 0.657 & \textbf{0.757} & 0.630 & 0.3617 & 75.20 & 3.82 & 38.3
\tabularnewline
  
5 & \textbf{0.863} & \textbf{0.859} & 0.678 & \textbf{0.2038} & \textbf{75.29} & {3.96} & 25.5
  & \textbf{0.658} & \textbf{0.757} & 0.626 & \textbf{0.3615} & 75.20 & 3.83 & \textbf{38.2}
  \tabularnewline

\bottomrule
\end{tabular}}
\vspace{-0.5em}

\end{table}

As shown in Table~\ref{tab:multiref}, the identity preservation
will improve as we introduce more reference faces.
The effect saturates at roughly five images, which is
similar to the observation in~\cite{refldm}.
Note, the IDS(REF) is calculated using the first available reference face.
That means the additional reference faces could pull the model
output slightly further from the first reference through Eq.~\eqref{eq:multicfg}.
Thus, IDS(REF) may drop with additional reference faces.
Nevertheless, our worst IDS(REF) is still higher than
previous methods in Table~\ref{tab:tab1}.

As shown in Figure~\ref{fig:fig1} for FFHQ-Ref Moderate, when the input LQ image contains a moderate degradation, the IDS performance gap among
the models is not very large in Table~\ref{tab:tab1}, hence it is highly recommended to zoom-in to
visually distinguish the differences in restored face details.
%
%
For instance, the black moles are well preserved
on the sixth row in Figure~\ref{fig:fig1}.
While other methods tends to excessively smooth the skin
texture, our model generates more realistic textures.
%

As shown in Figure~\ref{fig:fig2} for FFHQ-Ref Severe,
when the LQ face is almost unrecognizable, our method can still
sufficiently leverage the reference face and generate a face that is very close
to the ground truth, preserving identity.
In contrast, almost every other method generates a visually different person in most cases, which justifies the consistent improvements
on the identity metrics of our method.

\begin{wraptable}{r}{0.55\linewidth}
\vspace{-2em}
\caption{Comparison with the state-of-the-art reference-based methods on the CelebA-Ref-Test~\cite{refldm} dataset.}
\label{tab:tab1-celeb}
\resizebox{\linewidth}{!}{
\setlength{\tabcolsep}{2pt}%
\begin{tabular}{cc|cccccc}
\toprule 
\multirow{2}{*}{\textbf{Method}} & \multirow{2}{*}{\textbf{\#REF}} & \multicolumn{6}{c}{\textbf{CelebA-Ref-Test}}\tabularnewline
\cmidrule{3-8}
 &  & IDS$\uparrow$ & FaceNet$\uparrow$ & IDS(REF)$\uparrow$ & LPIPS$\downarrow$ & MUSIQ$\uparrow$ & NIQE$\downarrow$ \tabularnewline
\midrule
RefLDM & 1 & 0.768 & 0.821 & 0.564 & 0.2453 & 72.11 & 4.75  \tabularnewline
RestorerID & 1 & 0.756 & 0.820 & 0.527 & 0.2690 & 74.86 & 5.22  \tabularnewline
\midrule 
Ours & 1 & \textbf{0.779} & \textbf{0.827} & \textbf{0.691} & \textbf{0.2310} &
\textbf{75.64} & \textbf{3.98} \tabularnewline
\bottomrule
\end{tabular}}
\vspace{-1em}

\end{wraptable}

As demonstrated in Table~\ref{tab:tab1-celeb} for CelebA-Ref-Test, our model still
achieves the best identity preservation compared to other reference-based
methods.
All the above experimental results demonstrate the effectiveness
of our method, especially in terms of identity preservation.

%

%
%

\subsection{Ablation Study and Discussions}
\label{sec:42}


\begin{table}[t]

\caption{Ablation study on the Composite Context (CC) and Hard Example Identity Loss (HID).}
\label{tab:abl1}

\vspace{-0.5em}
\resizebox{\linewidth}{!}{
\setlength{\tabcolsep}{2pt}%
\begin{tabular}{cc|ccccccc|ccccccc}
\toprule 
\multicolumn{2}{c|}{\textbf{Modules}} & \multicolumn{7}{c|}{\textbf{FFHQ-Ref Moderate}} & \multicolumn{7}{c}{\textbf{FFHQ-Ref Severe}}\tabularnewline
\midrule 
CC & HID & IDS$\uparrow$ & FaceNet$\uparrow$ & IDS(REF)$\uparrow$ & LPIPS$\downarrow$ & MUSIQ$\uparrow$ & NIQE$\downarrow$ & FID$\downarrow$ & IDS$\uparrow$ & FaceNet$\uparrow$ & IDS(REF)$\uparrow$ & LPIPS$\downarrow$ & MUSIQ$\uparrow$ & NIQE$\downarrow$ & FID$\downarrow$\tabularnewline
\midrule

- & - & 0.811 & 0.841 & 0.565 & 0.2104 & \textbf{76.02} & \textbf{3.85} & 26.0 &
0.231 & 0.637 & 0.168 & 0.3896 & 73.85 & \textbf{3.67} & 43.4\tabularnewline

$\checkmark$ & - &
0.822 & 0.847 & 0.584 & 0.2074 & 75.66 & 3.89 & 25.9 &
0.345 & 0.675 & 0.288 & 0.3694 & \textbf{75.46} & 3.83 & \textbf{38.0}\tabularnewline

$\checkmark$ & $\checkmark$ &
\textbf{0.843} & \textbf{0.850} & \textbf{0.732} & \textbf{0.2054} & 75.29 & {3.96} & \textbf{25.5} &
\textbf{0.609} & \textbf{0.743} & \textbf{0.712} & \textbf{0.3647} & 75.22 & {3.84} & {38.3}
\tabularnewline

\bottomrule
\end{tabular}}

\end{table}


\begin{table}[t]

\caption{Ablation study on individual components of Composite Context. The evaluation of different combinations is carried out by using different attention masks with the same model checkpoint.}
\label{tab:abl2}

\vspace{-0.5em}
\resizebox{\linewidth}{!}{
\setlength{\tabcolsep}{2pt}%
\begin{tabular}{cc|ccccccc|ccccccc}
\toprule 
\multicolumn{2}{c}{\textbf{Composite Context}} & \multicolumn{7}{|c|}{\textbf{FFHQ-Ref Moderate}} & \multicolumn{7}{c}{\textbf{FFHQ-Ref Severe}}\tabularnewline
\midrule 
High-Level & General & IDS$\uparrow$ & FaceNet$\uparrow$ & IDS(REF)$\uparrow$ & LPIPS$\downarrow$ & MUSIQ$\uparrow$ & NIQE$\downarrow$ & FID$\downarrow$ & IDS$\uparrow$ & FaceNet$\uparrow$ & IDS(REF)$\uparrow$ & LPIPS$\downarrow$ & MUSIQ$\uparrow$ & NIQE$\downarrow$ & FID$\downarrow$\tabularnewline
\midrule

- & - & 
0.738 & 0.805 & 0.516 & 0.2196 & 74.43 & 3.99 & 27.9 &
0.186 & 0.616 & 0.137 & 0.3875 & 73.03 & 3.92 & 47.4 \tabularnewline

- & $\checkmark$ & 
0.770 & 0.821 & 0.567 & 0.2087 & 75.01 & \textbf{3.94} & 25.7 &
0.348 & 0.666 & 0.320 & 0.3713 & 74.51 & \textbf{3.83} & 40.1 \tabularnewline

$\checkmark$ & - & 
0.835 & 0.846 & 0.707 & 0.2094 & 75.20 & 3.98 & 25.9 &
0.535 & 0.717 & 0.625 & 0.3800 & 74.95 & 3.85 & 40.2 \tabularnewline

$\checkmark$ & $\checkmark$ &
\textbf{0.843} & \textbf{0.850} & \textbf{0.732} & \textbf{0.2054} & \textbf{75.29} & {3.96} & \textbf{25.5} &
\textbf{0.609} & \textbf{0.743} & \textbf{0.712} & \textbf{0.3647} & \textbf{75.22} & {3.84} & \textbf{38.3}
\tabularnewline

\bottomrule
\end{tabular}}

\end{table}


\begin{table}[t]

\caption{Ablation Study on Individual Components of the Hard Identity Loss.}
\label{tab:abl3}

\vspace{-0.5em}
\resizebox{\linewidth}{!}{
\setlength{\tabcolsep}{3pt}%
\begin{tabular}{c|ccccccc|ccccccc}
\toprule 
\multirow{2}{*}{$\lambda$} & \multicolumn{7}{c|}{\textbf{FFHQ-Ref Moderate}} & \multicolumn{7}{c}{\textbf{FFHQ-Ref Severe}}\tabularnewline
\cmidrule{2-15}
 & IDS$\uparrow$ & FaceNet$\uparrow$ & IDS(REF)$\uparrow$ & LPIPS$\downarrow$ & MUSIQ$\uparrow$ & NIQE$\downarrow$ & FID$\downarrow$ & IDS$\uparrow$ & FaceNet$\uparrow$ & IDS(REF)$\uparrow$ & LPIPS$\downarrow$ & MUSIQ$\uparrow$ & NIQE$\downarrow$ & FID$\downarrow$\tabularnewline
\midrule

$0$ &
\textbf{0.844} & \textbf{0.855} & 0.621 & 0.2039 & 75.36 & 3.97 & 25.5 &
0.485 & 0.712 & 0.465 & 0.3664 & 75.22 & 3.85 & 38.5
\tabularnewline

$0.6$ &
\uline{0.843} & \uline{0.850} & \uline{0.732} & {0.2054} & 75.29 & {3.96} & {25.5} &
\textbf{0.609} & \textbf{0.743} & \uline{0.712} & 0.3647 & 75.22 & {3.84} & {38.3}
\tabularnewline

$1$ &
0.779 & 0.821 & \textbf{0.794} & 0.2076 & 75.43 & 3.95 & 25.6 &
\uline{0.605} & \uline{0.742} & \textbf{0.768} & 0.3666 & 75.29 & 3.90 & 38.8
\tabularnewline
\bottomrule
\end{tabular}}

\end{table}

We conduct the ablation study in a hierarchical way,
firstly, coarse-grained based on
the two Composite Context and Hard Example Identity Loss modules.
Then we conduct the fine-grained ablation study for each component in these modules.

\textbf{Module-wise Ablation.}
Since the two modules are independent of each other, we conduct
the ablation study by removing some of them, and then retrain the model.
As shown in Table~\ref{tab:abl1},
both the context and loss contribute significantly to the final performance,
because the removal of any of them will lead to a major performance drop.
Removing both makes the model degenerate into a no-reference face
restoration model,
which lags behind our model too much in identity preservation.
This means both Composite Context and Hard Example Identity Loss are effective.
Next, we conduct an ablation study on the individual components of these
modules.

\textbf{Composite Context Ablation.}
As shown in Table~\ref{tab:abl2}, we study the contribution of individual
components in the Composite Context by applying attention masks
during inference.
It can be seen in the table that all the multi-level components, including high-level and general components
clearly contribute significantly to the final performance, as the removal
of any of them will lead to a performance drop, especially on the
FFHQ-Ref Severe test dataset.
%
%

\textbf{Hard Example Identity Loss Ablation.}
As shown in Table~\ref{tab:abl3}, we conduct an ablation study on the individual components
in the Hard Example Identity Loss, by adjusting the balancing parameter $\lambda$ 
between the HQ image and the reference image in Eq.~\eqref{eq:hid}.
According to the results, when we only use the ID loss with the HQ image ($\lambda=0$), the IDS(REF) is much lower, so is IDS on FFHQ-Ref Severe.
When we only use the ID loss with the REF image ($\lambda=1$), the IDS
will be traded off with IDS(REF).
Hence, we empirically set the $\lambda$ parameter as $0.6$
by default, by considering all the three identity preservation metrics.
%
The case where the Hard Identity Loss is removed ($w_\text{HID}=0$)
is at the second row of Table~\ref{tab:abl1}, and that
leads to a much lower performance regardless of the $\lambda$ parameter.

\begin{wrapfigure}{r}{0.45\linewidth}

\vspace{-1em}
\noindent\resizebox{1.0\linewidth}{!}{%
\setlength{\tabcolsep}{0.2mm}%
\renewcommand{\arraystretch}{0.5}%
\begin{tabular}{cccc}

\midrule
LQ & REF & Result & HQ \\
\midrule

\includegraphics[width=0.3\linewidth]{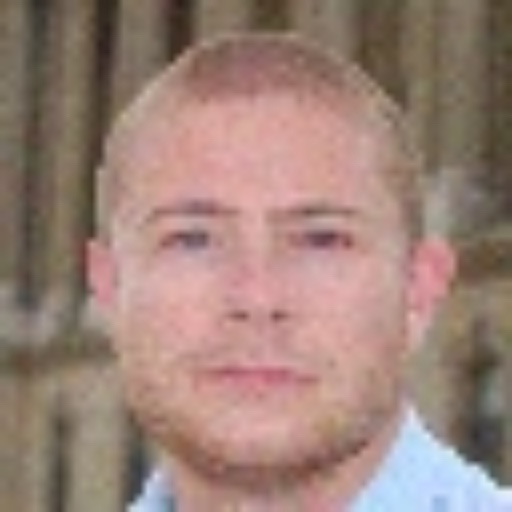} &
\multirow[c]{2}{*}[0.7cm]{%
\includegraphics[width=0.3\linewidth]{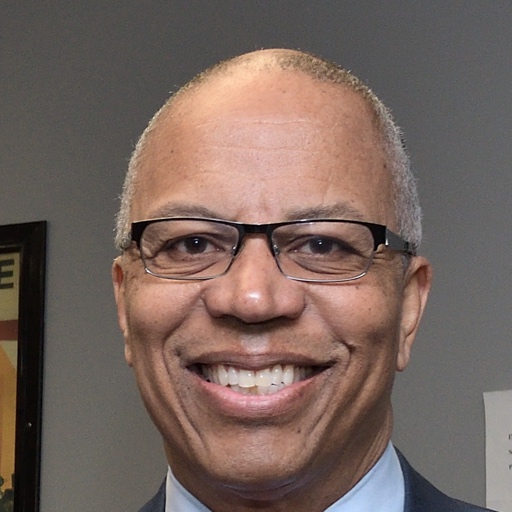}%
} &
\includegraphics[width=0.3\linewidth]{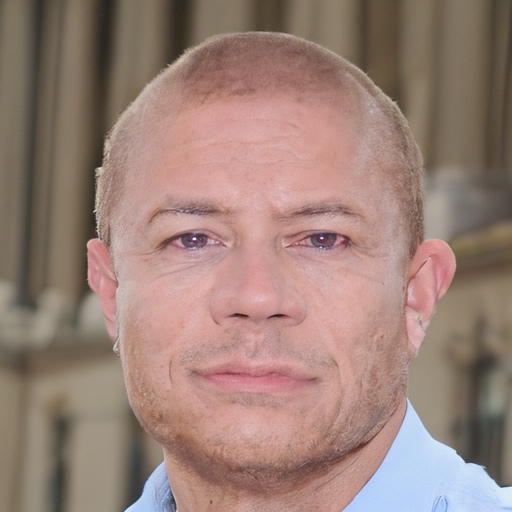} &
\multirow[c]{2}{*}[0.7cm]{%
\includegraphics[width=0.3\linewidth]{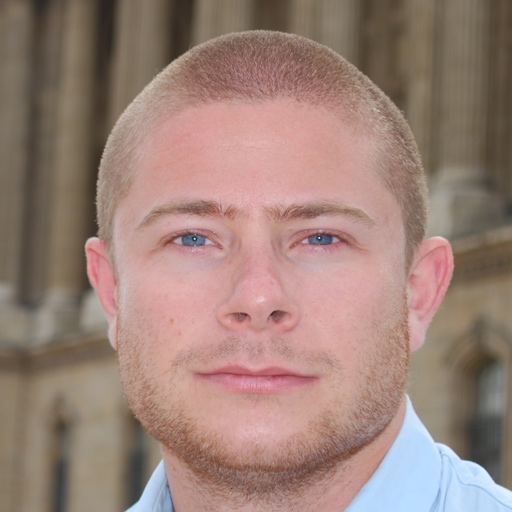}%
} \\

\includegraphics[width=0.3\linewidth]{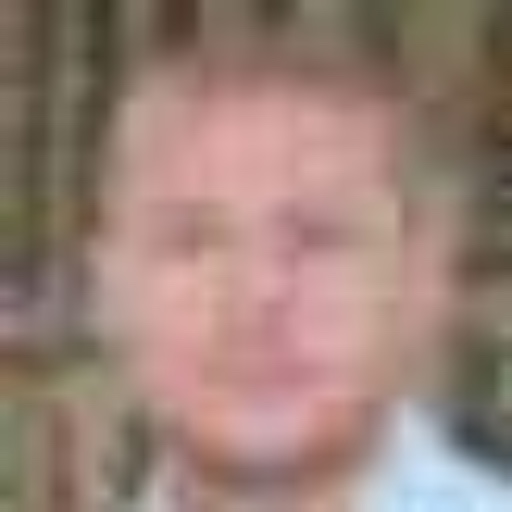} &
&
\includegraphics[width=0.3\linewidth]{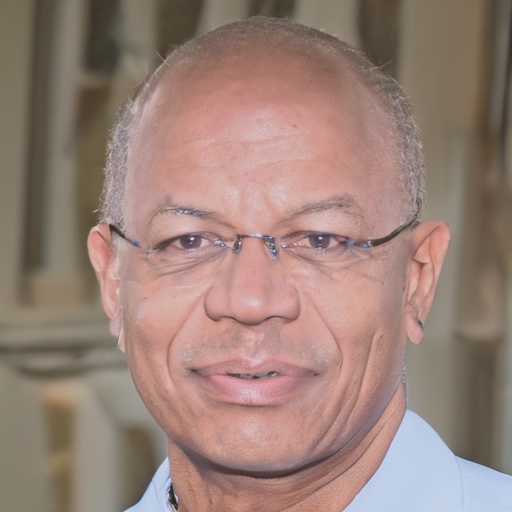} &
\\
\midrule

\end{tabular}}

\vspace{-0.5em}
\caption{Demonstration of the impact of reference face image, by deliberately
supplying the model with a reference face of a wrong identity.
The first row is from FFHQ-Ref Moderate, and the second row is from FFHQ-Ref Severe.}
\label{fig:wrongface}
\vspace{-0.7em}
\end{wrapfigure}

\textbf{Influence of Reference Face.}
The above ablation study supports the effectiveness of our method when using the a \textit{correct} reference face.
While the problem of reference-based face restoration assumes a reference face with
correct identity is provided, it is difficult to guarantee in real-world applications.
To demonstrate the influence of the reference face, we deliberately use a
\textit{wrong} reference
face, as shown in Figure~\ref{fig:wrongface}.
According to our observation, when the input LQ image has moderate information loss
with the person identity roughly recognizable, our model will largely follow the LQ, and
add slight identity-related details to the result, as shown in the first row in the figure.
When the input LQ has severe information loss with the person identity almost unrecognizable,
the REF face image becomes dominant and show stronger impact in the resulting image.

This phenomenon, on the one hand, further demonstrates the effectiveness of our method through
the influence of the reference face.
On the other hand, it also implies the importance of ensuring the correct identity in
real-world applications for reference-based face restoration.

\textbf{Limitations and Future Work.}
\textbf{(1)} The face restoration model is trained on simulated
degradation pipelines~\cite{realesrgan},
which means it may underperform on
in-the-wild face images with unknown degradations.
\textbf{(2)} While the reference-based
face restoration task assume high-quality reference images
are available, in practical scenarios the reference image quality may vary.
In that case, exploring different weighting mechanisms in
Eq.~\eqref{eq:multicfg} could be a potential direction for
future exploration.
\textbf{(3)}
A large-scale high-quality dataset
for reference-based face restoration is
still missing, and the FFHQ-Ref~\cite{refldm} training set only
contains $18816$ images.
One potential approach to obtain more data could be
filtering face recognition datasets~\cite{webface}.
We leave these directions for future study.

\noindent\textbf{Broader Impact.}
Our method aims at restoring low-quality face images,
with the goal of contributing positively to society.
However, being a diffusion-based face generative model, it might be abused to forge DeepFake~\cite{deepfake} images.
We suggest real-world face restoration service providers apply invisible watermarks~\cite{videoseal}
to the generated result to mitigate potential risks and
negative societal impact.

\section{Conclusion}

We present a reference-based face restoration method,
highlighting two key modules: Composite Context and Hard Example Identity Loss that
focus on identity preservation.
The two key modules are designed to better exploit reference face images,
while all the existing works leverage it to a lesser extent.
Meanwhile, the proposed method can be extended for the multi-reference case
in a training-free manner.
Experimental results on the FFHQ-Ref and CelebA-Ref-Test datasets demonstrate
the effectiveness of our proposed method.
Ablation studies on the Composite Context and Hard Example Identity Loss suggest that all the proposed modules in our method,
including the individual components in the modules, are effective and
make a considerable impact on identity preservation.
%
%

\textbf{Acknowledgement.}
We are grateful to Mojtaba Ardakani for his assistance with the diffusion formulation and code review, to Julian Iseringhausen for his help in fine-tuning the VAE on face images, and to Xiang Zhu for insightful discussions.

%


{
    \small
    \bibliography{arxiv}
}


\appendix


\section{Detailed Quantitative Results and More Visualizations}

%

\begin{figure}[t]

\noindent%
\setlength{\tabcolsep}{0.2mm}%
\renewcommand{\arraystretch}{0.4}%
\resizebox{\linewidth}{!}{%
\begin{tabular}{cccccccc}

\cellcolor{gray!20} LQ & \cellcolor{gray!20} REF & CodeFormer & DiffBIR & RefLDM & RestorerID & \cellcolor{yellow!50} \textbf{Ours} & \cellcolor{gray!20} HQ \\

\midrule

\suppfigmoderate{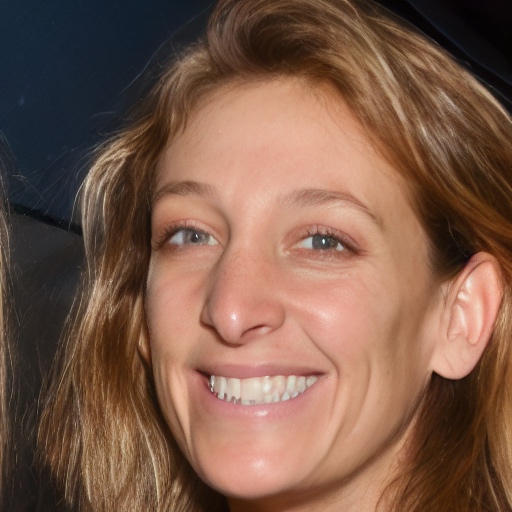} \\
\suppfigmoderate{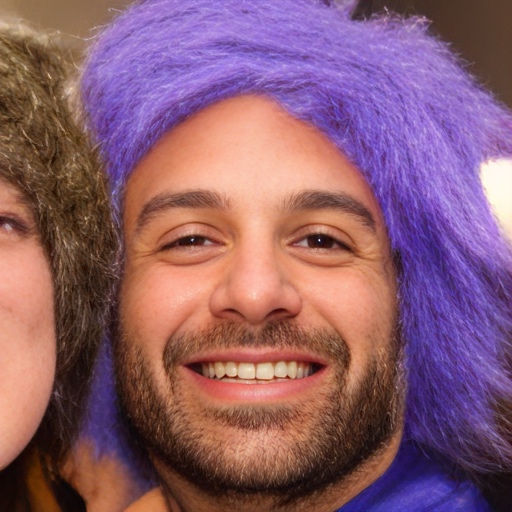} \\

\suppfigmoderate{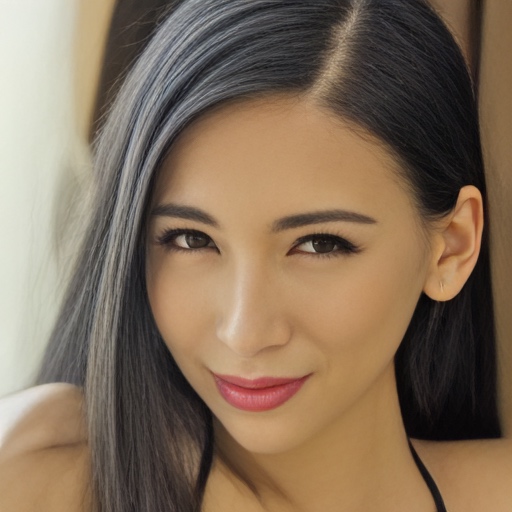} \\
\suppfigmoderate{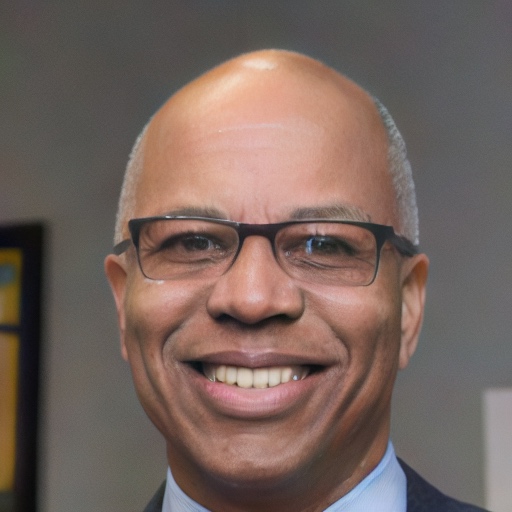} \\

\suppfigmoderate{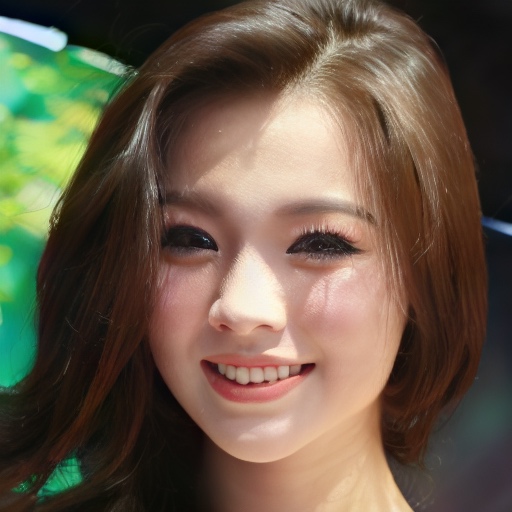} \\
\suppfigmoderate{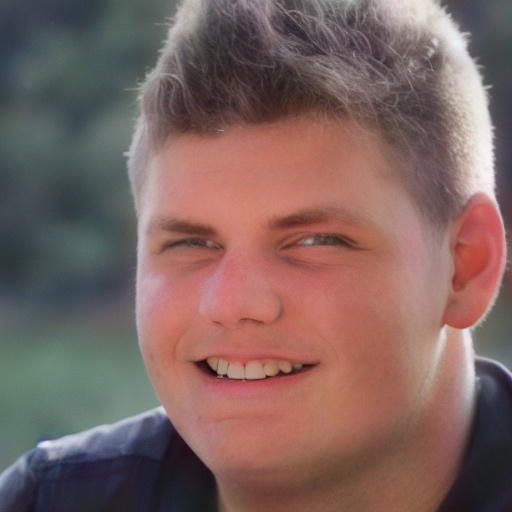} \\

\suppfigmoderate{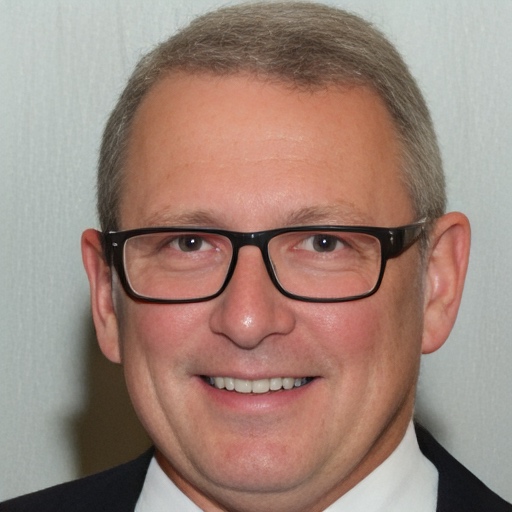} \\
\suppfigmoderate{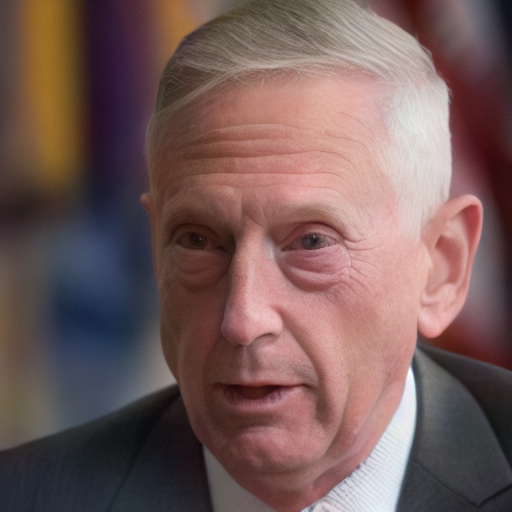} \\

\suppfigmoderate{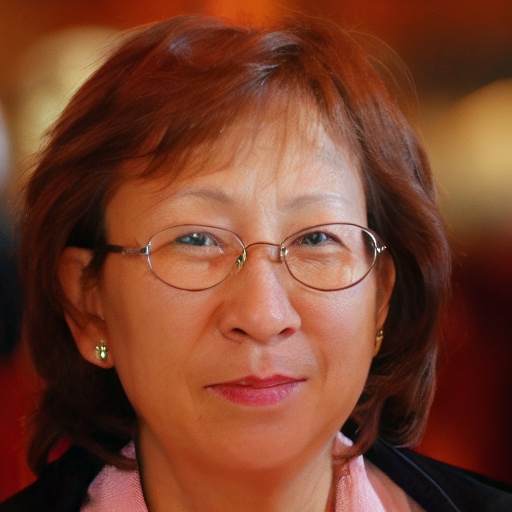} \\
\suppfigmoderate{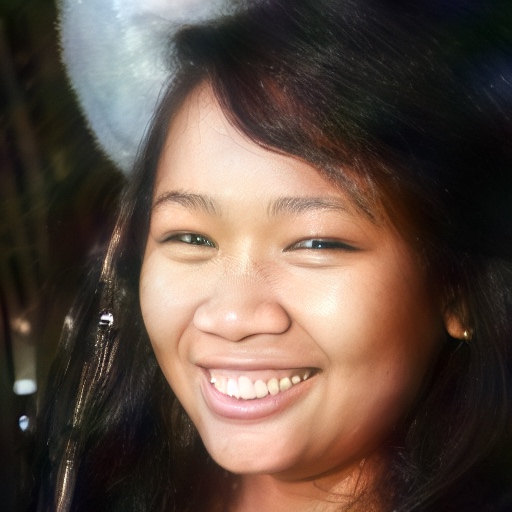} \\

\midrule

\cellcolor{gray!20} LQ & \cellcolor{gray!20} REF & CodeFormer & DiffBIR & RefLDM & RestorerID & \cellcolor{yellow!50} \textbf{Ours} & \cellcolor{gray!20} HQ \\

\end{tabular}}

\vspace{-0.3em}
\caption{Additional qualitative comparison with other state-of-the-art face restoration methods on FFHQ-Ref Moderate test set. The ``REF'' column is the high-quality reference face image.}
\label{fig:supp-vis-moderate}
\end{figure}

\begin{figure}[t]

\noindent%
\setlength{\tabcolsep}{0.2mm}%
\renewcommand{\arraystretch}{0.4}%
\resizebox{\linewidth}{!}{%
\begin{tabular}{cccccccc}

\cellcolor{gray!20} LQ & \cellcolor{gray!20} REF & CodeFormer & DiffBIR & RefLDM & RestorerID & \cellcolor{yellow!50} \textbf{Ours} & \cellcolor{gray!20} HQ \\

\midrule

\suppfigsevere{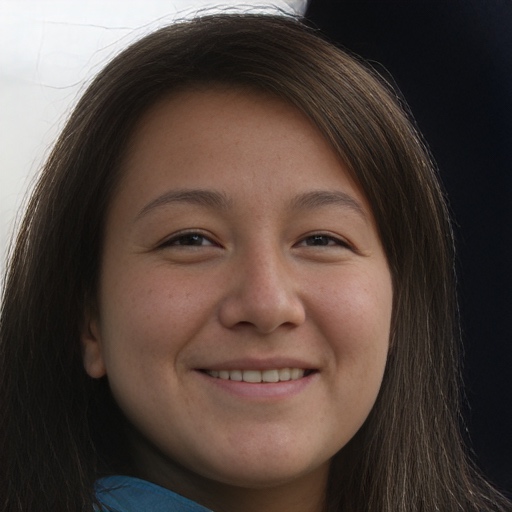} \\

\suppfigsevere{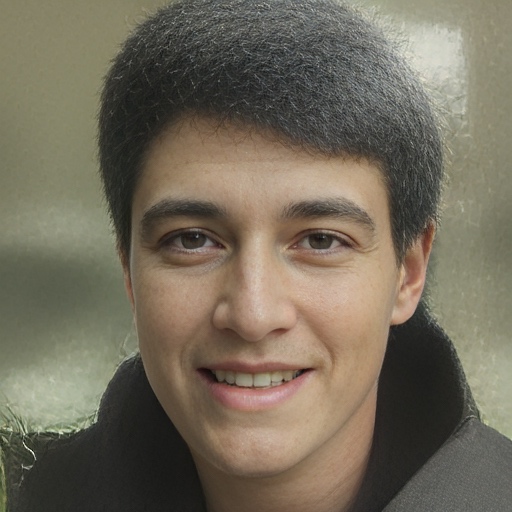} \\

\suppfigsevere{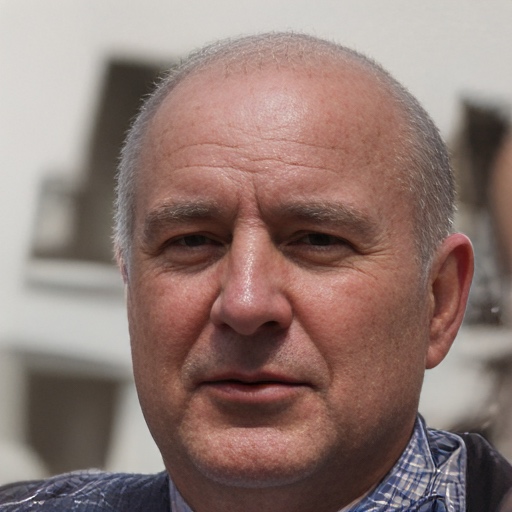} \\
\suppfigsevere{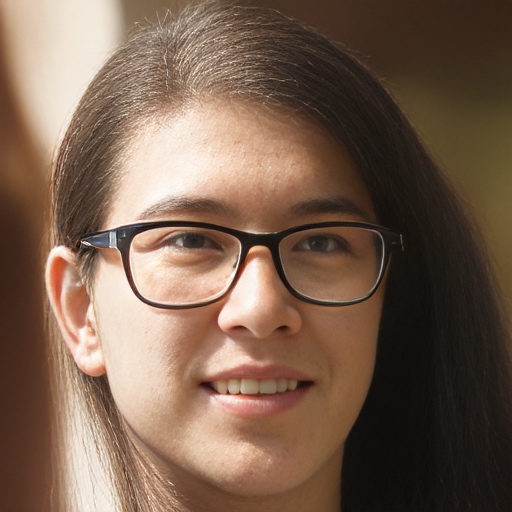} \\

\suppfigsevere{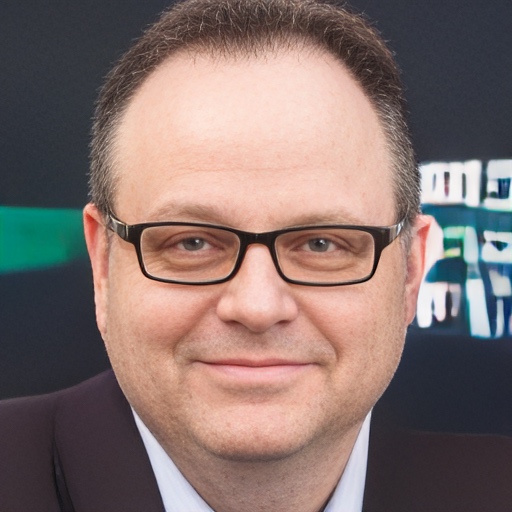} \\
\suppfigsevere{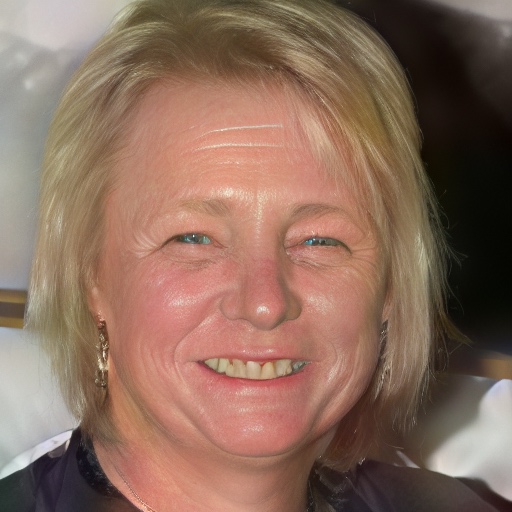} \\

\suppfigsevere{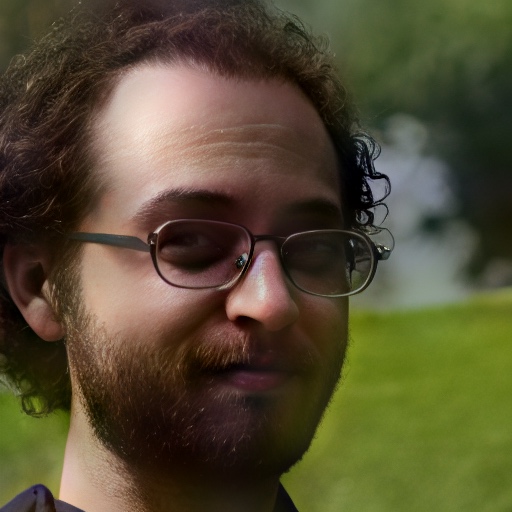} \\
\suppfigsevere{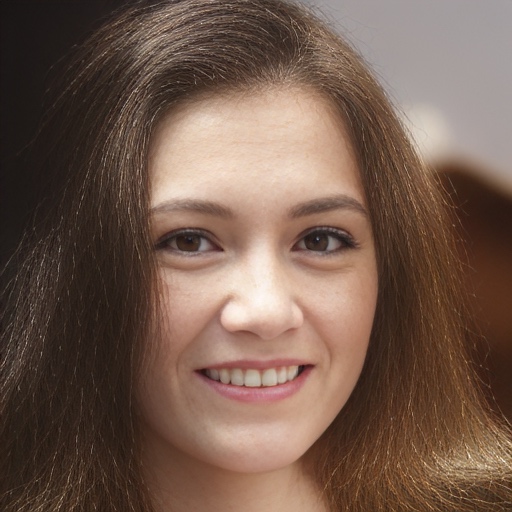} \\

\suppfigsevere{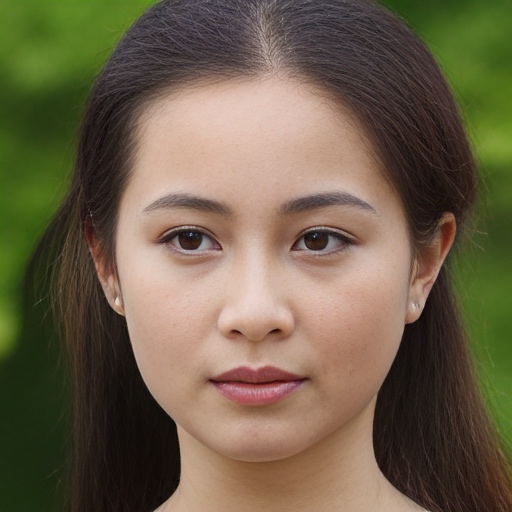} \\
\suppfigsevere{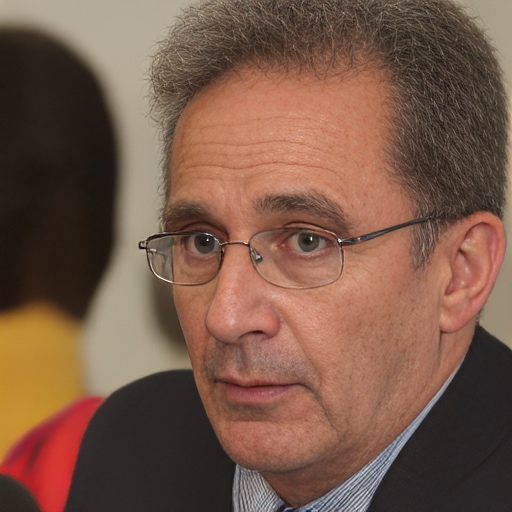} \\

\midrule

\cellcolor{gray!20} LQ & \cellcolor{gray!20} REF & CodeFormer & DiffBIR & RefLDM & RestorerID & \cellcolor{yellow!50} \textbf{Ours} & \cellcolor{gray!20} HQ \\

\end{tabular}}

\vspace{-0.3em}
\caption{Additional qualitative comparison with other state-of-the-art face restoration methods on FFHQ-Ref Severe test set. The ``REF'' column is the high-quality reference face image.}
\label{fig:supp-vis-severe}
\end{figure}

\begin{figure}[t]

\noindent%
\setlength{\tabcolsep}{0.2mm}%
\renewcommand{\arraystretch}{0.4}%
\resizebox{\linewidth}{!}{%
\begin{tabular}{cccccccc}

\cellcolor{gray!20} LQ & \cellcolor{gray!20} REF & CodeFormer & DiffBIR & RefLDM & RestorerID & \cellcolor{yellow!50} \textbf{Ours} & \cellcolor{gray!20} HQ \\

\midrule

\suppfigceleba{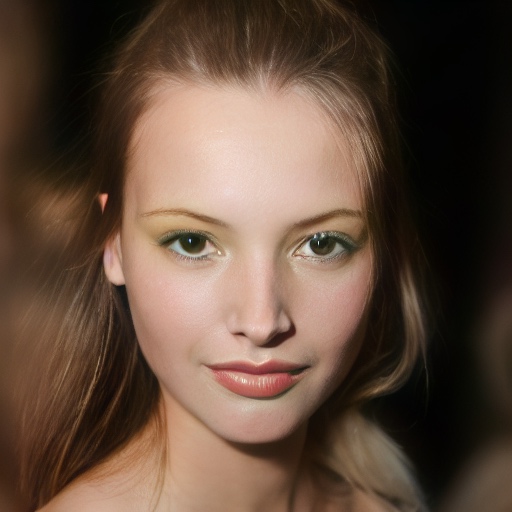} \\
\suppfigceleba{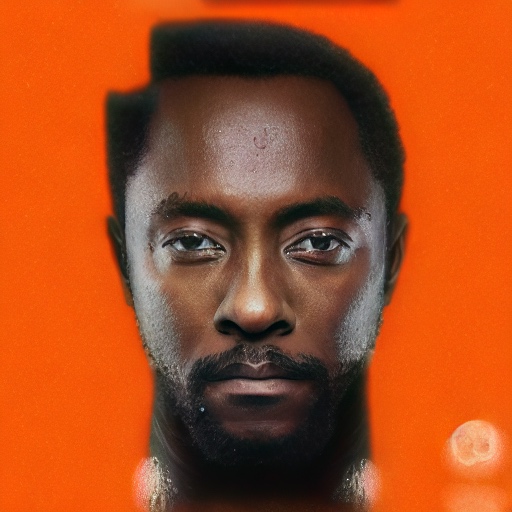} \\

\suppfigceleba{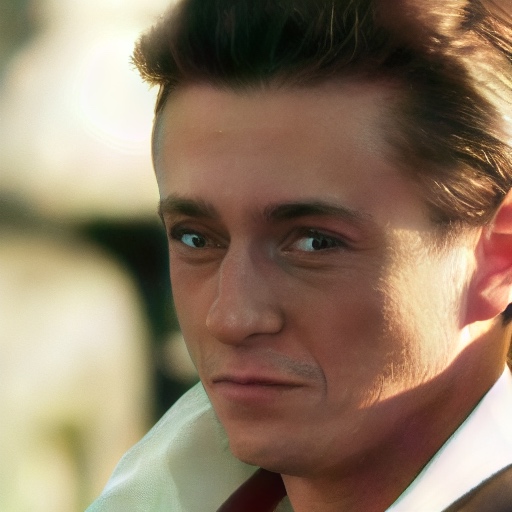} \\
\suppfigceleba{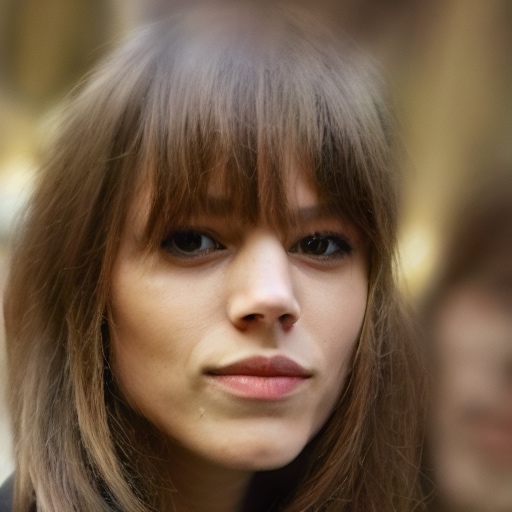} \\

\suppfigceleba{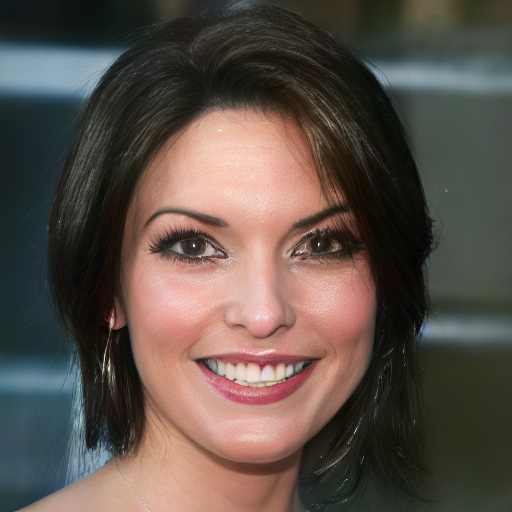} \\
\suppfigceleba{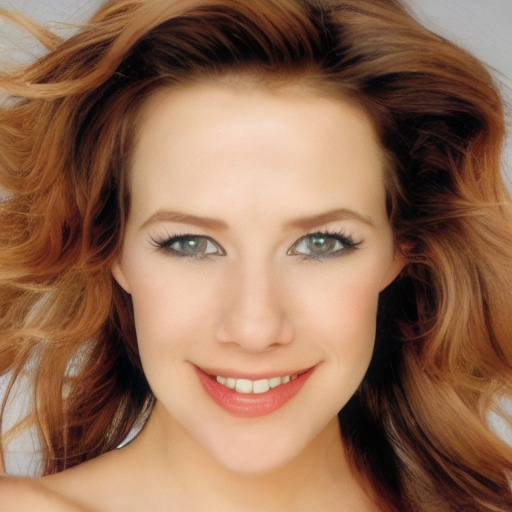} \\

\midrule

\cellcolor{gray!20} LQ & \cellcolor{gray!20} REF & CodeFormer & DiffBIR & RefLDM & RestorerID & \cellcolor{yellow!50} \textbf{Ours} & \cellcolor{gray!20} HQ \\

\end{tabular}}

\vspace{-0.3em}
\caption{Additional qualitative comparison with other state-of-the-art face restoration methods on CelebA-Ref-Test test set. The ``REF'' column is the high-quality reference face image.}
\label{fig:supp-vis-celeba}
\end{figure}

The detailed results and comparison with state-of-the-art methods on FFHQ-Ref Moderate,
FFHQ-Ref Severe, and CelebA-Ref-Test can be found in Table~\ref{tab:moderate-full},
Table~\ref{tab:severe-full}, and Table~\ref{tab:celeba-full}, respectively.
Additional visualizations are provided in Figure~\ref{fig:supp-vis-moderate},
Figure~\ref{fig:supp-vis-severe}, and Figure~\ref{fig:supp-vis-celeba}
for those three test sets.

\begin{figure}[h]
\resizebox{\linewidth}{!}{%
\setlength{\tabcolsep}{1pt}%
\begin{tabular}{cccc}
\toprule
LQ & \#REF=1 & \#REF=5 & HQ\\
\midrule
\includegraphics[width=0.2\linewidth]{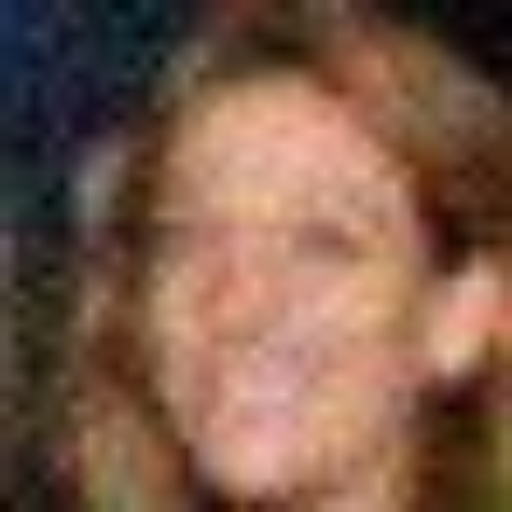}
& \includegraphics[width=0.2\linewidth]{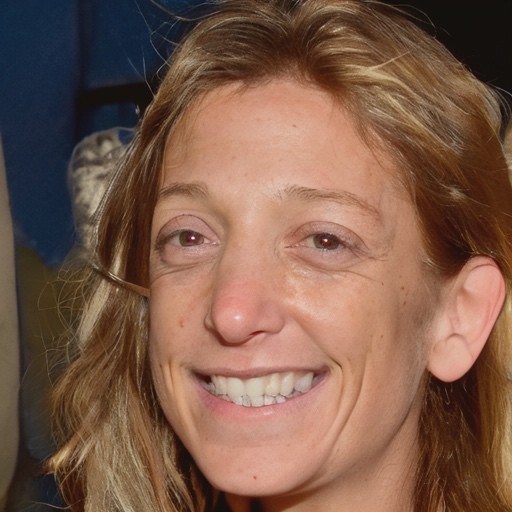}
& \includegraphics[width=0.2\linewidth]{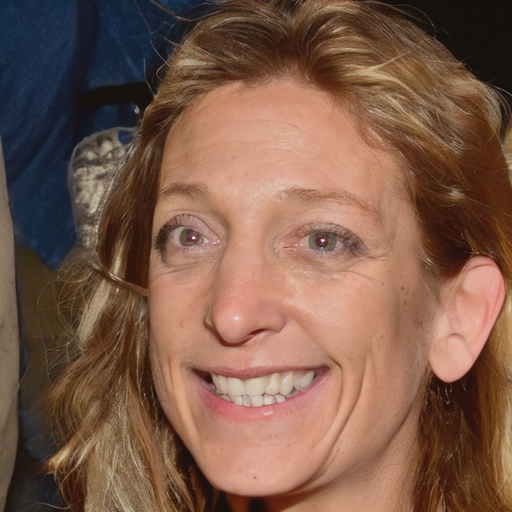}
& \includegraphics[width=0.2\linewidth]{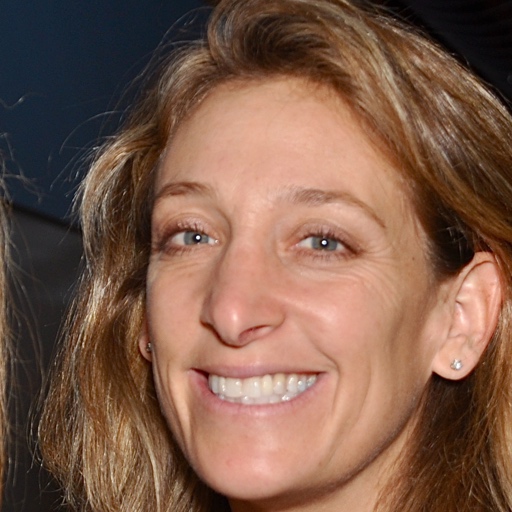}
\\
\includegraphics[width=0.2\linewidth]{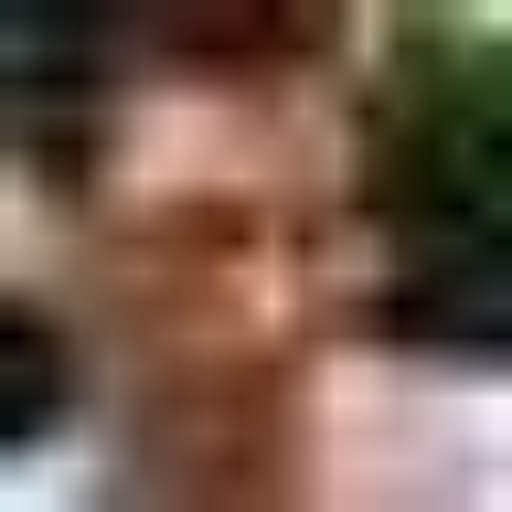}
& \includegraphics[width=0.2\linewidth]{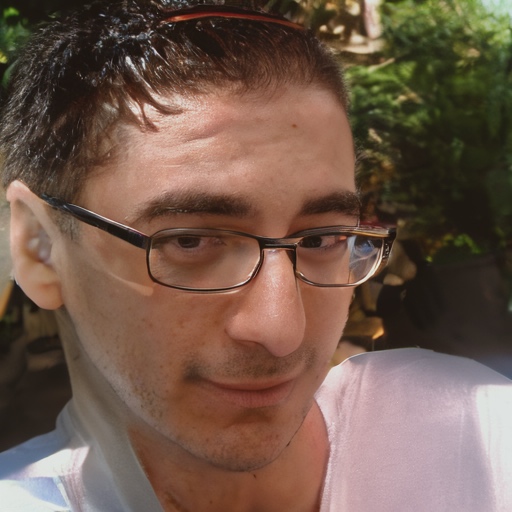}
& \includegraphics[width=0.2\linewidth]{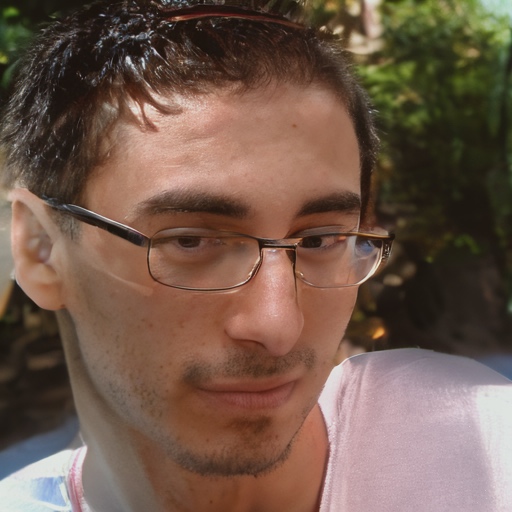}
& \includegraphics[width=0.2\linewidth]{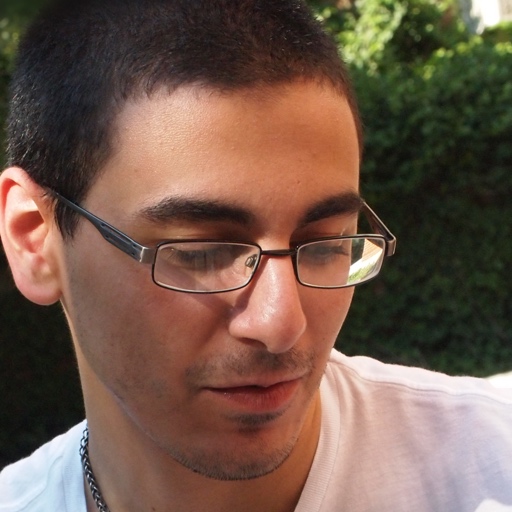}
\\
\includegraphics[width=0.2\linewidth]{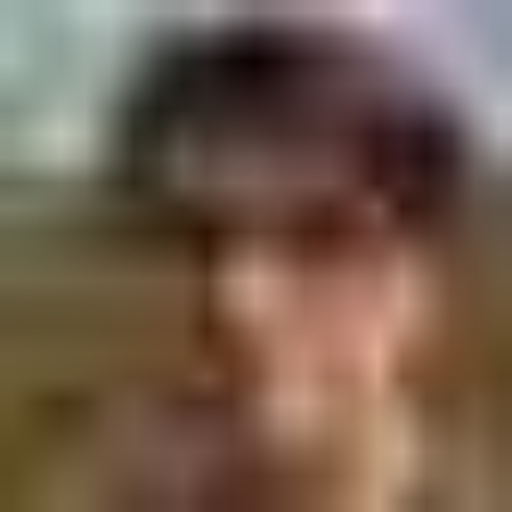}
& \includegraphics[width=0.2\linewidth]{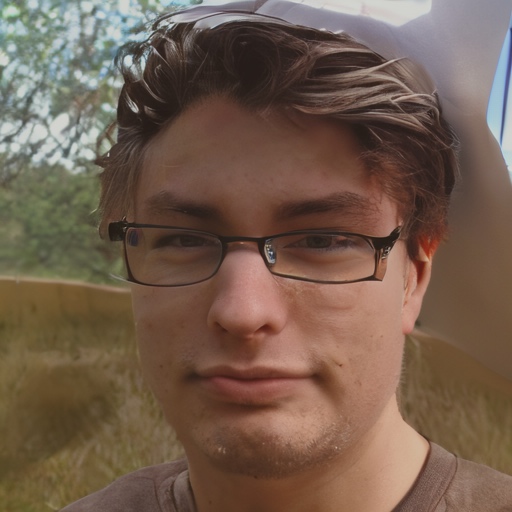}
& \includegraphics[width=0.2\linewidth]{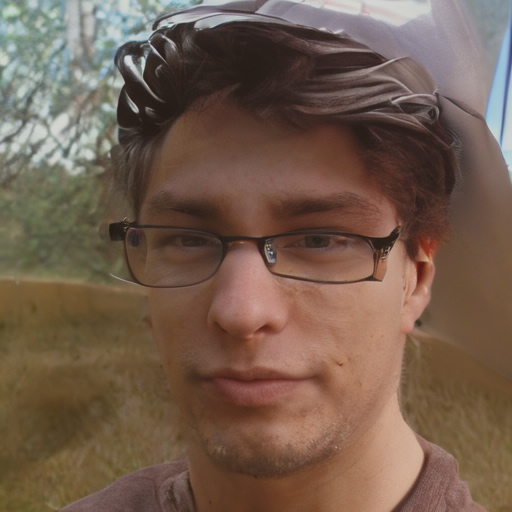}
& \includegraphics[width=0.2\linewidth]{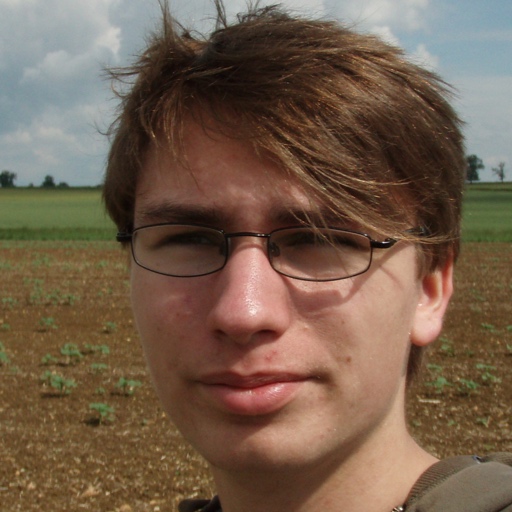}
\\
\bottomrule
\end{tabular}}
\caption{Visualization of multi-reference face restoration on FFHQ-Ref Severe.}
\label{fig:supp-multiref}
\end{figure}

The detailed multi-reference face restoration results and comparison with state-of-the-art methods
can be found in Table~\ref{tab:moderate-multi-full},
Table~\ref{tab:severe-multi-full}, and Table~\ref{tab:celeba-multi-full}, respectively for
the three test sets.
Some visualizations are provided in Figure~\ref{fig:supp-multiref} to demonstrate
the effect of using more than one reference faces.

\begin{figure}[h]
\resizebox{\linewidth}{!}{%
\setlength{\tabcolsep}{1pt}%
\begin{tabular}{cccc}
\toprule
LQ & REF & Result & HQ \\
\midrule
\includegraphics[width=0.2\linewidth]{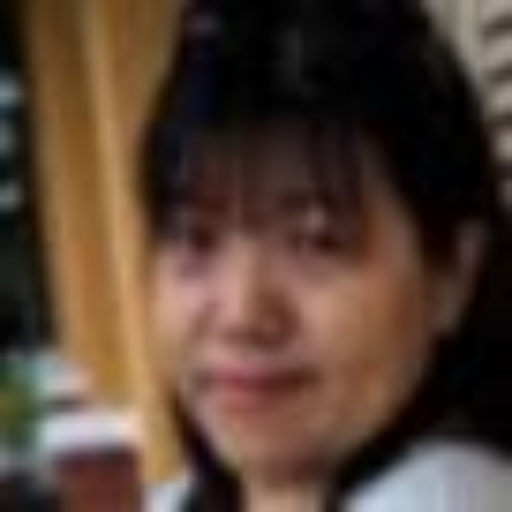}
& \includegraphics[width=0.2\linewidth]{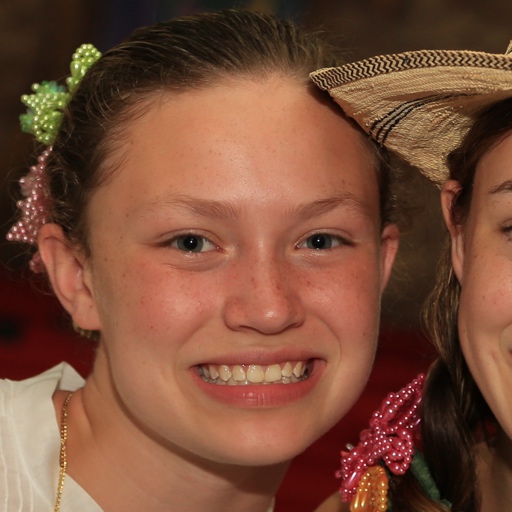}
& \includegraphics[width=0.2\linewidth]{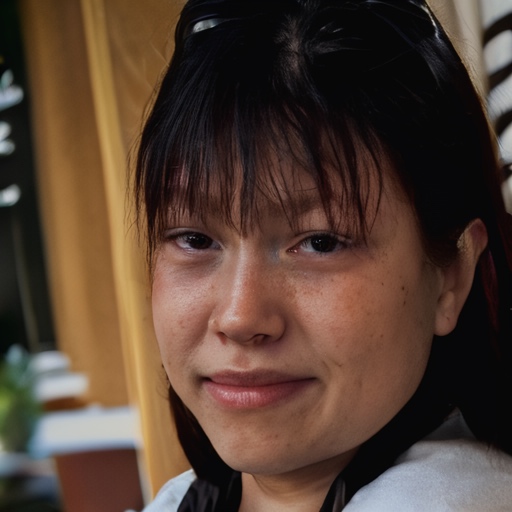}
& \includegraphics[width=0.2\linewidth]{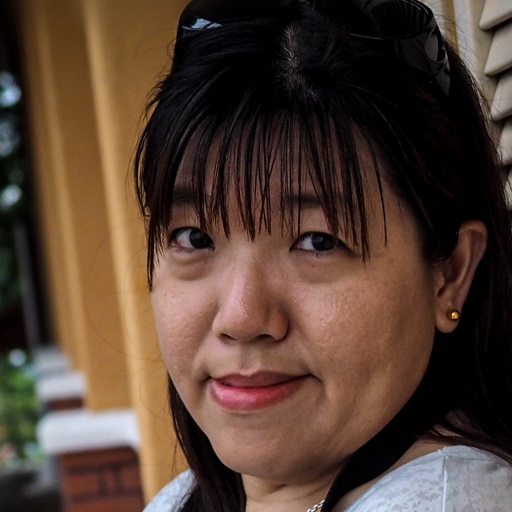}
\\
\includegraphics[width=0.2\linewidth]{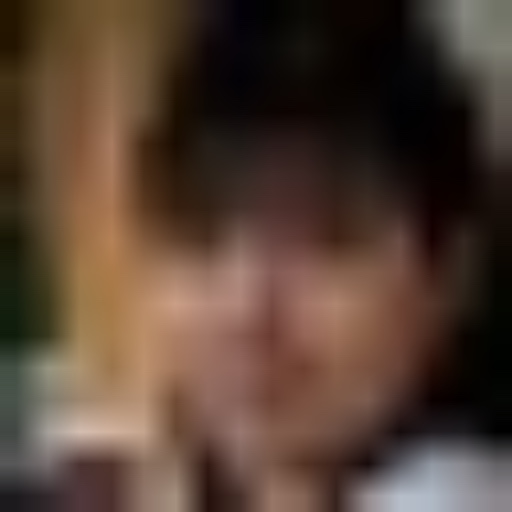}
& \includegraphics[width=0.2\linewidth]{supp-wrongface/62148-ref1.jpg}
& \includegraphics[width=0.2\linewidth]{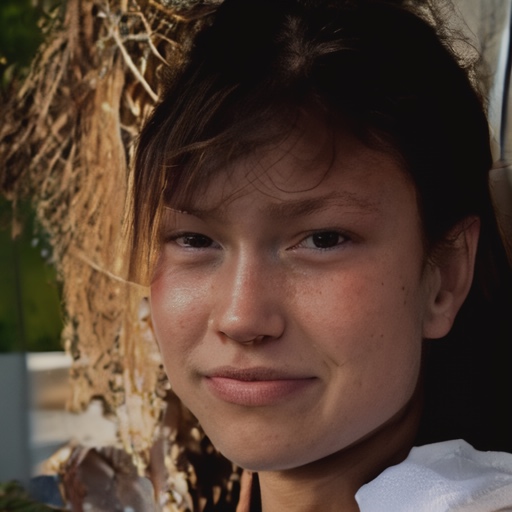}
& \includegraphics[width=0.2\linewidth]{supp-wrongface/62148-hq.jpg}
\\
\midrule
\includegraphics[width=0.2\linewidth]{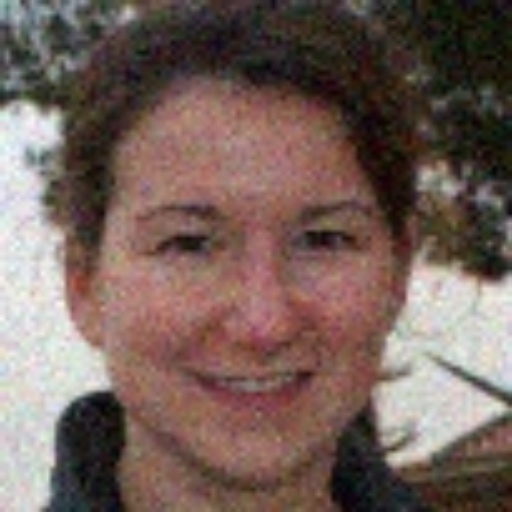}
& \includegraphics[width=0.2\linewidth]{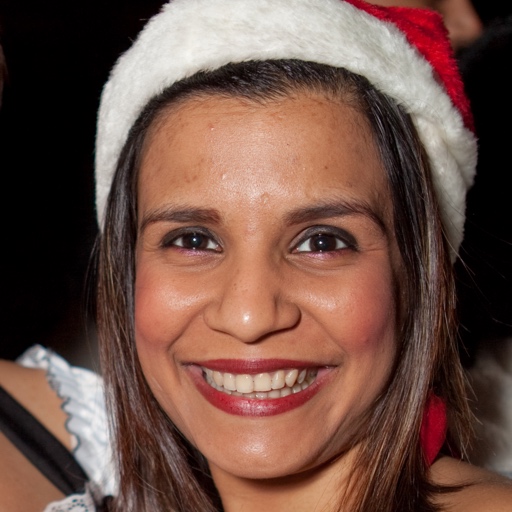}
& \includegraphics[width=0.2\linewidth]{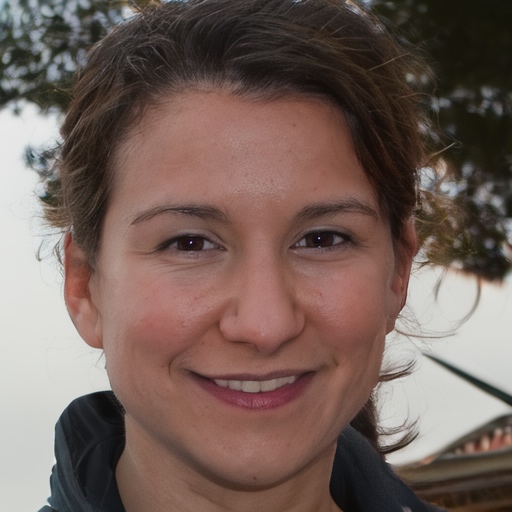}
& \includegraphics[width=0.2\linewidth]{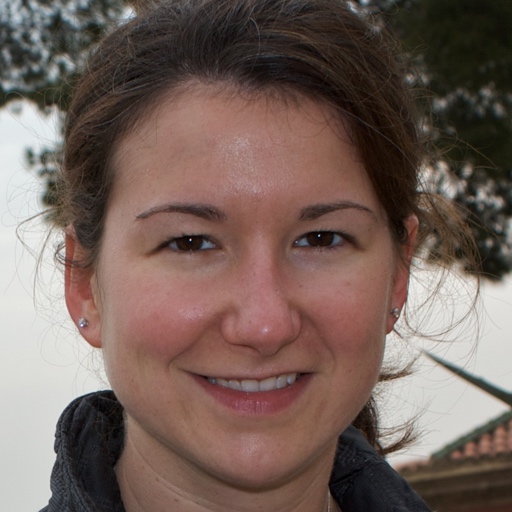}
\\
\includegraphics[width=0.2\linewidth]{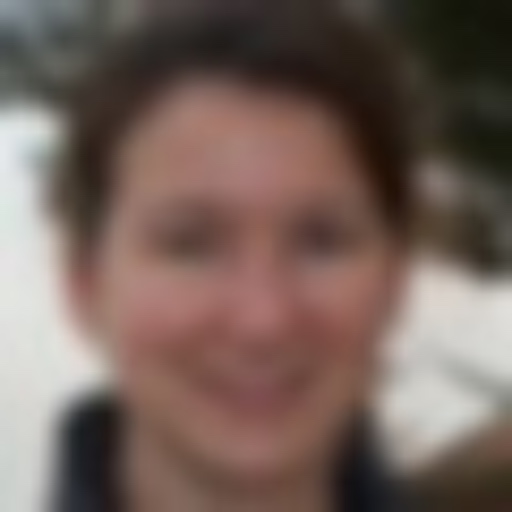}
& \includegraphics[width=0.2\linewidth]{supp-wrongface/01590-ref1.jpg}
& \includegraphics[width=0.2\linewidth]{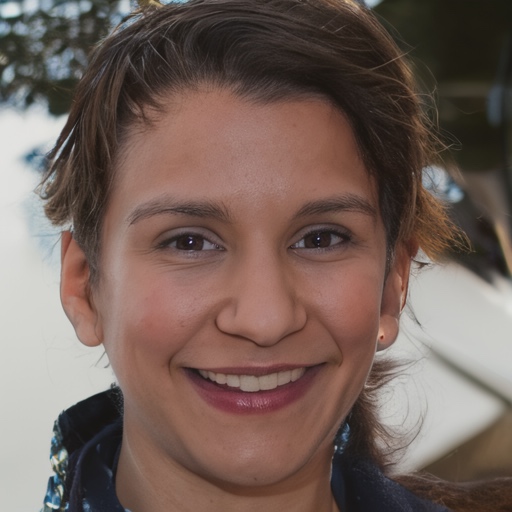}
& \includegraphics[width=0.2\linewidth]{supp-wrongface/01590-hq.jpg}
\\
\bottomrule
\end{tabular}}
\caption{Additional visualizations with wrong reference face. This table is a continuation of the Figure~5 in the manuscript. As discussed in the manuscript, a wrong reference face will leads to some ``identity blending'' effect depending on how much information is lost from the low-quality input face.}
\label{fig:wrong-ff}
\end{figure}

\begin{figure}[h]
\resizebox{\linewidth}{!}{%
\setlength{\tabcolsep}{1pt}%
\begin{tabular}{cccc}
\toprule
LQ & REF & Result & HQ \\
\midrule
\includegraphics[width=0.2\linewidth]{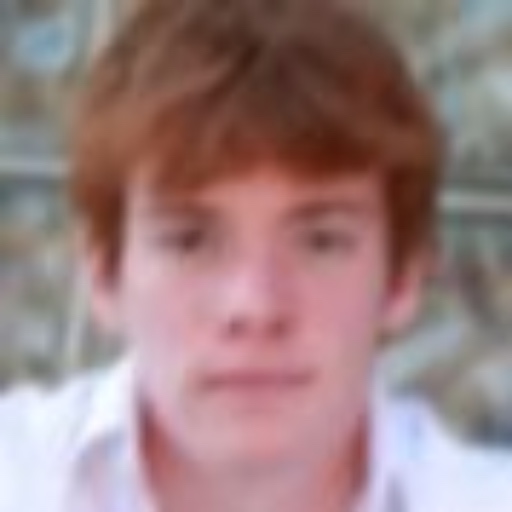}
& \includegraphics[width=0.2\linewidth]{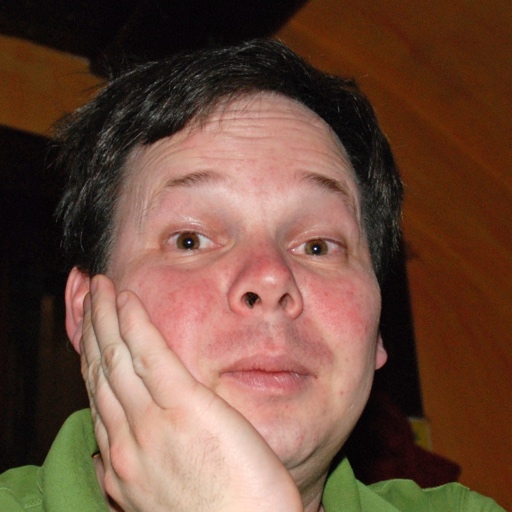}
& \includegraphics[width=0.2\linewidth]{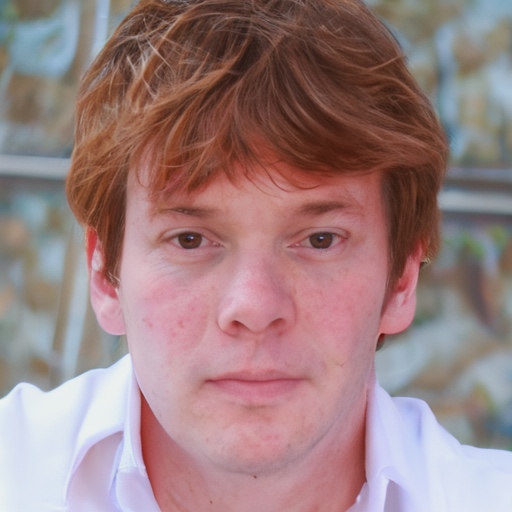}
& \includegraphics[width=0.2\linewidth]{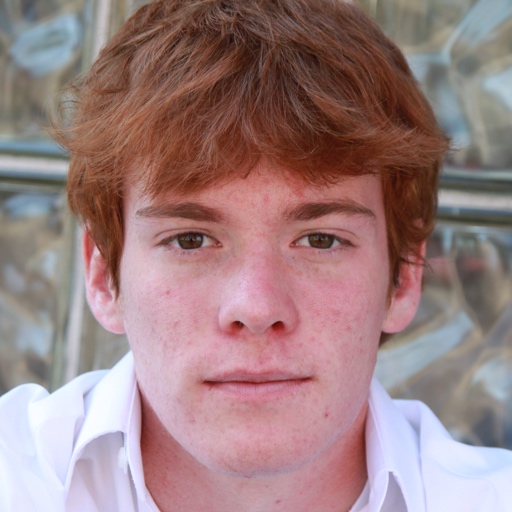}
\\
\includegraphics[width=0.2\linewidth]{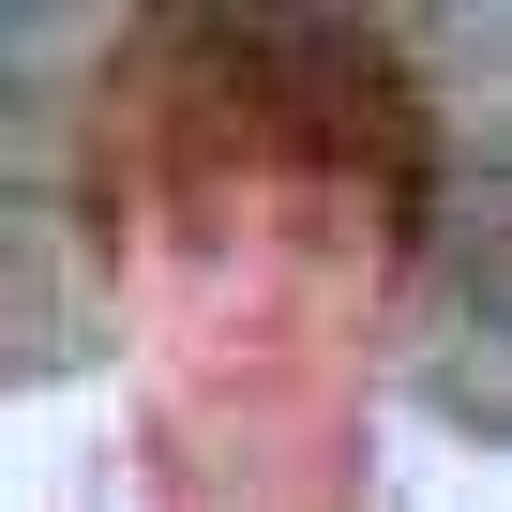}
& \includegraphics[width=0.2\linewidth]{supp-wrongface/06010-ref1.jpg}
& \includegraphics[width=0.2\linewidth]{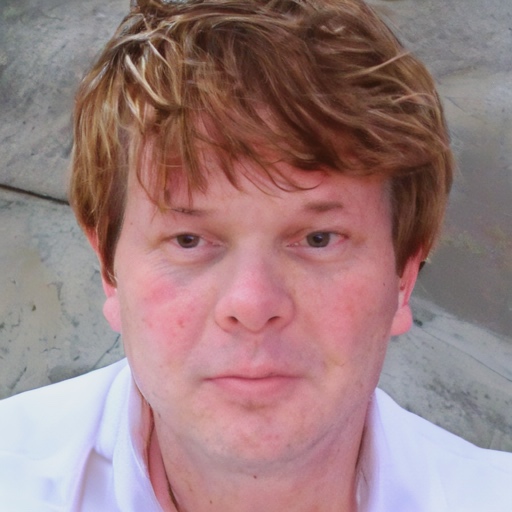}
& \includegraphics[width=0.2\linewidth]{supp-wrongface/06010-hq.jpg}
\\
\midrule
\includegraphics[width=0.2\linewidth]{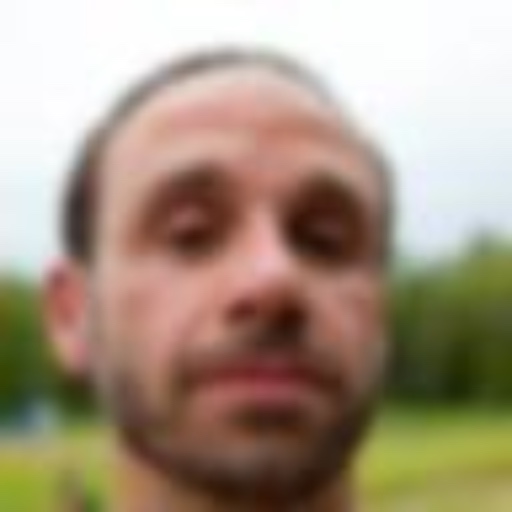}
& \includegraphics[width=0.2\linewidth]{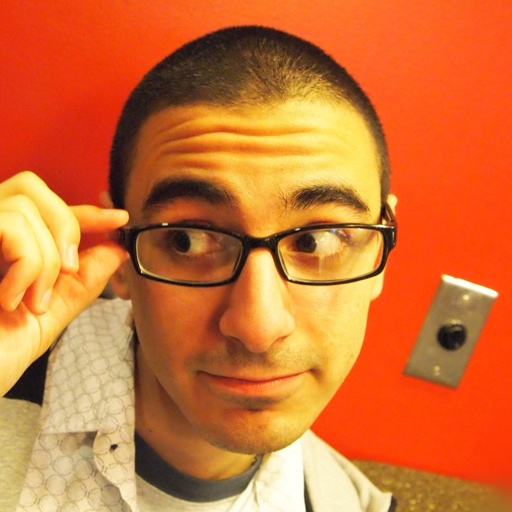}
& \includegraphics[width=0.2\linewidth]{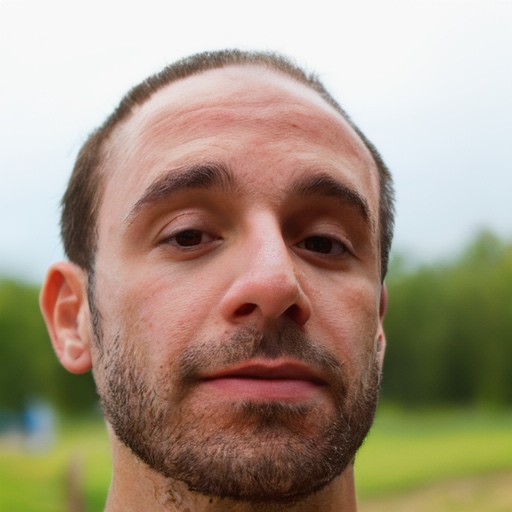}
& \includegraphics[width=0.2\linewidth]{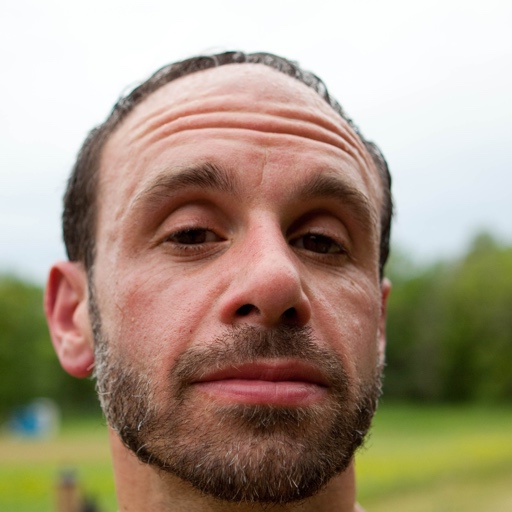}
\\
\includegraphics[width=0.2\linewidth]{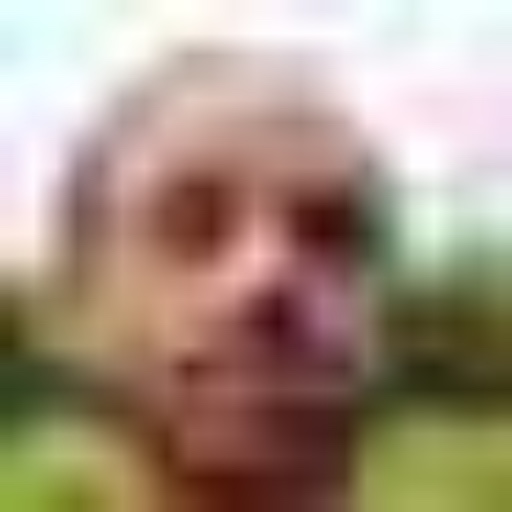}
& \includegraphics[width=0.2\linewidth]{supp-wrongface/12912-ref1.jpg}
& \includegraphics[width=0.2\linewidth]{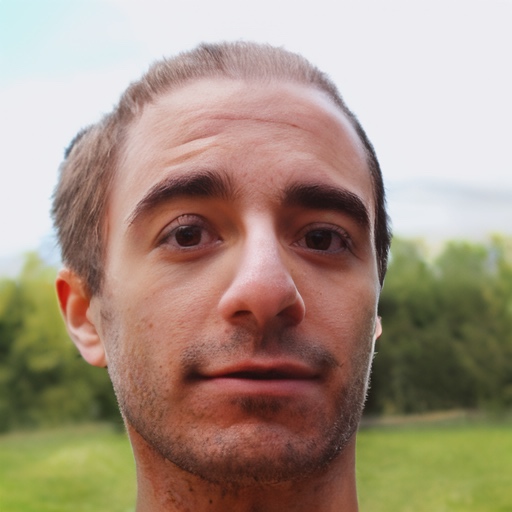}
& \includegraphics[width=0.2\linewidth]{supp-wrongface/12912-hq.jpg}
\\
\bottomrule
\end{tabular}}
\caption{Additional visualizations with wrong reference face. This table is a continuation of the Figure~5 in the manuscript. As discussed in the manuscript, a wrong reference face will leads to some ``identity blending'' effect depending on how much information is lost from the low-quality input face.}
\label{fig:wrong-mm}
\end{figure}

\begin{figure}[h]
\resizebox{\linewidth}{!}{%
\setlength{\tabcolsep}{1pt}%
\begin{tabular}{cccc}
\toprule
LQ & REF & Result & HQ \\
\midrule
\includegraphics[width=0.2\linewidth]{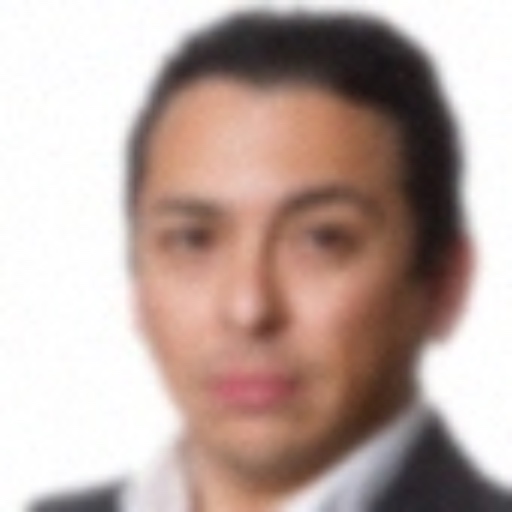}
& \includegraphics[width=0.2\linewidth]{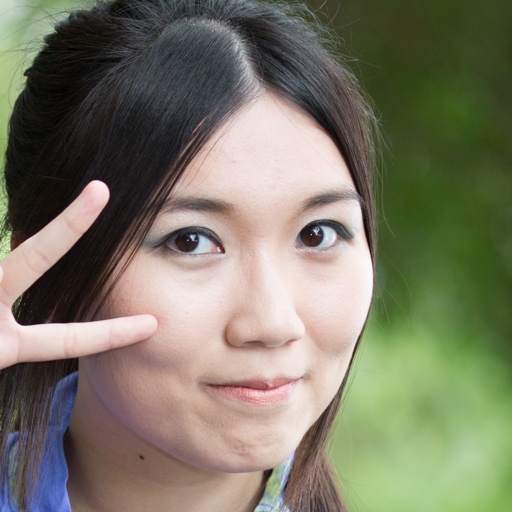}
& \includegraphics[width=0.2\linewidth]{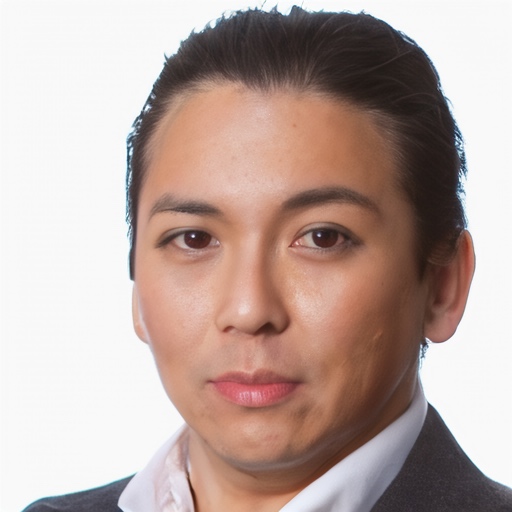}
& \includegraphics[width=0.2\linewidth]{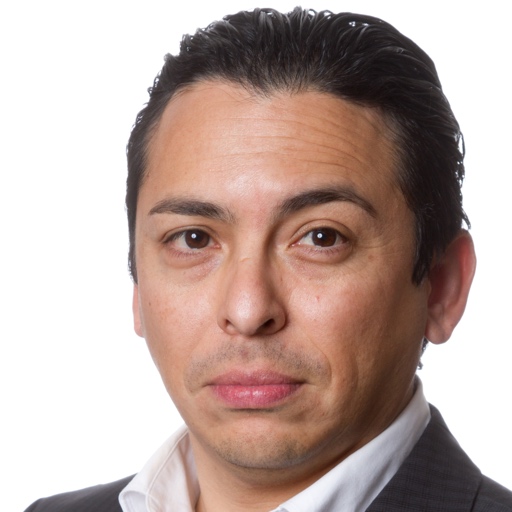}
\\
\includegraphics[width=0.2\linewidth]{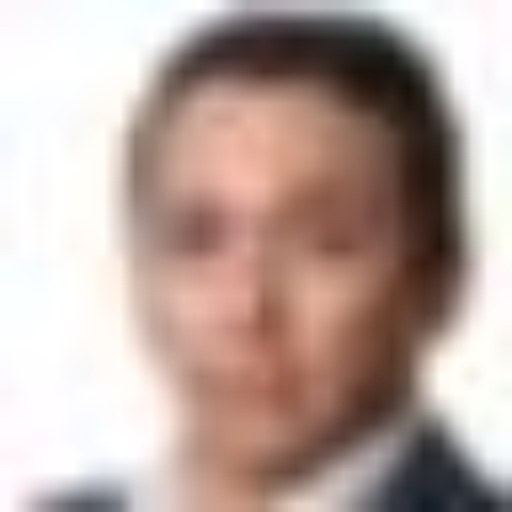}
& \includegraphics[width=0.2\linewidth]{supp-wrongface/62384-ref1.jpg}
& \includegraphics[width=0.2\linewidth]{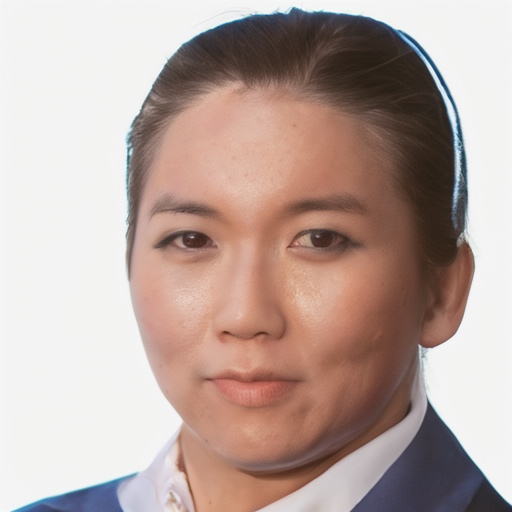}
& \includegraphics[width=0.2\linewidth]{supp-wrongface/62384-hq.jpg}
\\
\midrule
\includegraphics[width=0.2\linewidth]{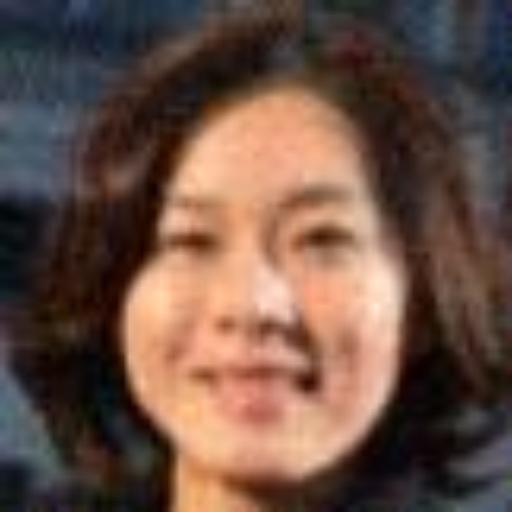}
& \includegraphics[width=0.2\linewidth]{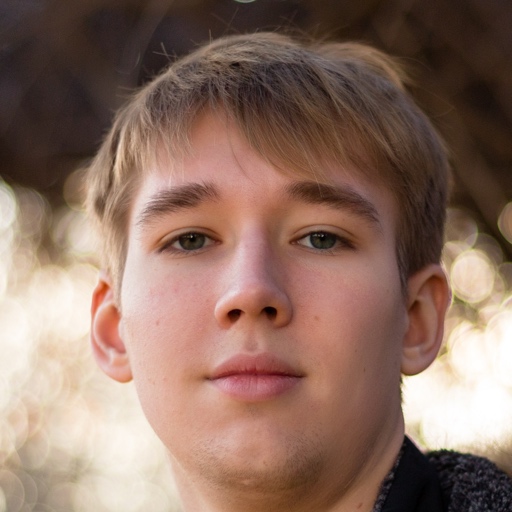}
& \includegraphics[width=0.2\linewidth]{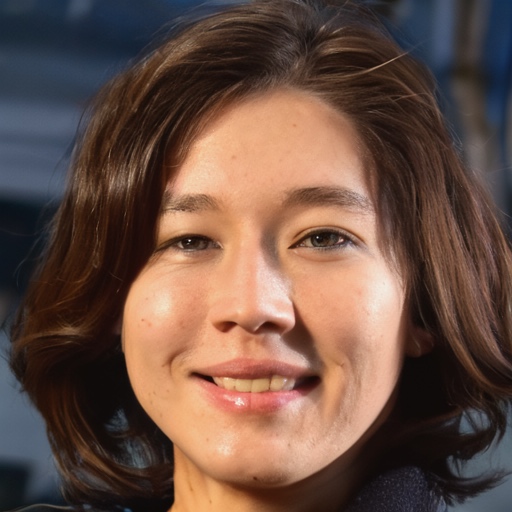}
& \includegraphics[width=0.2\linewidth]{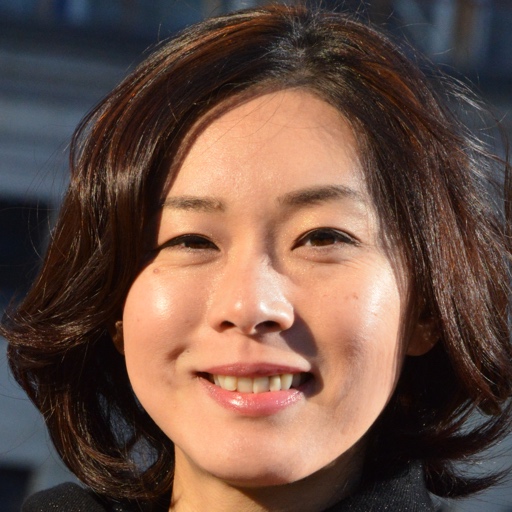}
\\
\includegraphics[width=0.2\linewidth]{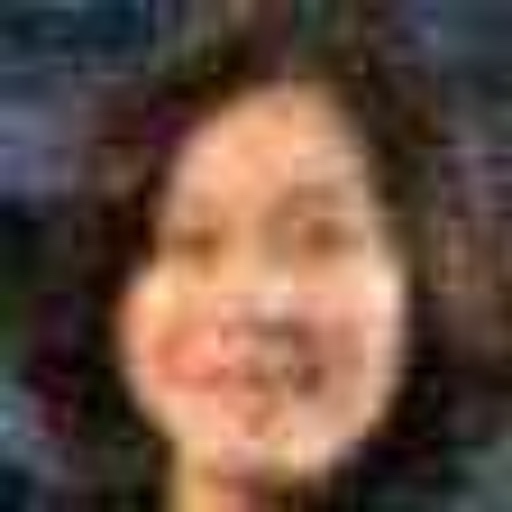}
& \includegraphics[width=0.2\linewidth]{supp-wrongface/08326-ref1.jpg}
& \includegraphics[width=0.2\linewidth]{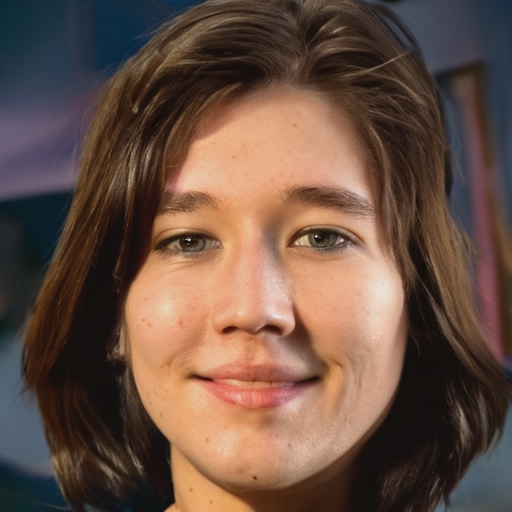}
& \includegraphics[width=0.2\linewidth]{supp-wrongface/08326-hq.jpg}
\\
\bottomrule
\end{tabular}}
\caption{Additional visualizations with wrong reference face. This table is a continuation of the Figure~5 in the manuscript. As discussed in the manuscript, a wrong reference face will leads to some ``identity blending'' effect depending on how much information is lost from the low-quality input face.}
\label{fig:wrong-mf}
\end{figure}

Additional visualizations with wrong reference face (as discussed in Section~4.2)
can be found in Figure~\ref{fig:wrong-ff}, Figure~\ref{fig:wrong-mm}, and Figure~\ref{fig:wrong-mf}.
A minor ablation study on the classifier-free guidance scale parameter can be found in Table~\ref{tab:supp-cfg}.

\begin{table}[h]
\caption{Detailed Quantitative Results on FFHQ-Ref Moderate test set.}
\label{tab:moderate-full}
\resizebox{\linewidth}{!}{%
\begin{tabular}{lc|ccccccc}
\toprule
\multirow{2}{*}{Method} & \multirow{2}{*}{\#REF} & \multicolumn{7}{c}{FFHQ-Ref Moderate}\\
\cmidrule{3-9}
                        &   & IDS & FaceNet & IDS(REF) & LPIPS & MUSIQ & NIQE & FID \\
\midrule
CodeFormer              & 0 & $0.783\pm 0.082$ & $0.822\pm 0.047$ & $0.545\pm 0.106$ & $0.1839\pm 0.0471$ & $75.88\pm 2.01$ & $4.38\pm 0.69$ & 31.7\\
DiffBIR                 & 0 & $0.831\pm 0.095$ & $0.842\pm 0.056$ & $0.575\pm 0.108$ & $0.2268\pm 0.0633$ & $76.64\pm 1.64$ & $5.72\pm 1.23$ & 34.9\\
\midrule
RefLDM                  & 1 & $0.826\pm 0.077$ & $0.837\pm 0.048$ & $0.624\pm 0.096$ &  $0.2210\pm 0.0583$ & $72.30\pm 4.89$ & $4.61\pm 0.64$ & 28.0\\
RestorerID              & 1 & $0.804\pm 0.099$ & $0.832\pm 0.054$ & $0.591\pm 0.096$ & $0.2350\pm 0.0688$ & $73.35\pm 5.12$ & $4.98\pm 0.81$ & 31.0\\
\midrule
Ours                    & 1 & $0.843\pm 0.076$  &  $0.850\pm 0.051$  & $0.732\pm 0.069$ & $0.2054\pm 0.0606$ & $75.29\pm 2.77$ & $3.96\pm 0.71$ & 25.5 \\
\bottomrule
\end{tabular}}
\end{table}

\begin{table}[h]
\caption{Detailed Quantitative Results on FFHQ-Ref Severe test set.}
\label{tab:severe-full}
\resizebox{\linewidth}{!}{%
\begin{tabular}{lc|ccccccc}
\toprule
\multirow{2}{*}{Method} & \multirow{2}{*}{\#REF} & \multicolumn{7}{c}{FFHQ-Ref Severe}                 \\
\cmidrule{3-9}
                        &   & IDS & FaceNet & IDS(REF) & LPIPS & MUSIQ & NIQE & FID \\
\midrule
CodeFormer              & 0 & $0.370\pm 0.150$ & $0.677\pm 0.061$ & $0.265\pm 0.132$ & $0.3113\pm 0.0801$ & $76.12\pm 1.94$ & $4.30\pm 0.70$ & 49.6\\
DiffBIR                 & 0 & $0.356\pm 0.144$ & $0.672\pm 0.058$ & $0.253\pm 0.124$ & $0.3606\pm 0.0879$ & $75.71\pm 2.81$ & $6.24\pm 1.22$ & 55.3\\
\midrule
RefLDM                  & 1 & $0.571\pm 0.110$ & $0.733\pm 0.052$ & $0.554\pm 0.112$ & $0.3366\pm 0.0756$ & $74.32\pm 3.36$ & $4.52\pm 0.62$ & 36.0\\
RestorerID              & 1 & $0.411\pm 0.110$ & $0.690\pm 0.052$ & $0.408\pm 0.103$ & $0.4130\pm 0.0741$ & $74.49\pm 3.41$ & $4.71\pm 0.65$ & 52.7\\
\midrule
Ours                    & 1 & $0.609\pm 0.089$ & $0.743\pm 0.048$ & $0.712\pm 0.068$ & $0.3647\pm 0.0722$ & $75.22\pm 2.46$ & $3.84\pm 0.64$ & 38.3\\
\bottomrule
\end{tabular}}
\end{table}

\begin{table}[h]
\caption{Detailed Quantitative Results on CelebA-Ref-Test test set.}
\label{tab:celeba-full}
\resizebox{\linewidth}{!}{%
\begin{tabular}{lc|ccccccc}
\toprule
\multirow{2}{*}{Method} & \multirow{2}{*}{\#REF} & \multicolumn{7}{c}{CelebA-Ref-Test}                 \\
\cmidrule{3-9}
                        &   & IDS & FaceNet & IDS(REF) & LPIPS & MUSIQ & NIQE & FID \\
\midrule
RefLDM                  & 1 & $0.768\pm 0.085$ & $0.821\pm 0.046$ & $0.564\pm 0.096$ & $0.2453\pm 0.0550$ & $72.11\pm 4.59$ & $4.75\pm 0.55$ & 19.4\\
RestorerID              & 1 & $0.756\pm 0.098$ & $0.820\pm 0.049$ & $0.527\pm 0.090$ & $0.2690\pm 0.0629$ & $74.86\pm 3.82$ & $5.22\pm 0.76$ & 25.4\\
\midrule
Ours                    & 1 & $0.779\pm 0.086$ & $0.827\pm 0.048$ & $0.691\pm 0.064$ & $0.2310\pm 0.0540$ & $75.64\pm 2.44$ & $3.98\pm 0.53$ & 18.4\\
\bottomrule
\end{tabular}}
\end{table}

\begin{table}[h]
\caption{Detailed Quantitative Results on FFHQ-Ref Moderate test set. Note, our multi-reference face support is training-free, while RefLDM's is not.}
\label{tab:moderate-multi-full}
\resizebox{\linewidth}{!}{%
\begin{tabular}{lc|ccccccc}
\toprule
\multirow{2}{*}{Method} & \multirow{2}{*}{\#REF} & \multicolumn{7}{c}{FFHQ-Ref Moderate}                 \\
\cmidrule{3-9}
                        &                        & IDS & FaceNet & IDS(REF) & LPIPS & MUSIQ & NIQE & FID \\
\midrule
RefLDM                  & 1 & $0.826\pm 0.077$ & $0.837\pm 0.048$ & $0.624\pm 0.096$ &  $0.2210\pm 0.0583$ & $72.30\pm 4.89$ & $4.61\pm 0.64$ & 28.0\\
Ours                    & 1 & $0.843\pm 0.076$  &  $0.850\pm 0.051$  & $0.732\pm 0.069$ & $0.2054\pm 0.0606$ & $75.29\pm 2.77$ & $3.96\pm 0.71$ & 25.5 \\
\midrule
RefLDM                  & 2                      & $0.839\pm 0.067$ & $0.844\pm 0.045$ & $0.630\pm 0.094$ & $0.2150\pm 0.0577$ & $73.25\pm 4.34$ & $4.57\pm 0.62$ & 27.6\\
Ours                    & 2                      & $0.857\pm 0.069$ & $0.856\pm 0.049$ & $0.693\pm 0.075$ & $0.2042\pm 0.0603$ & $75.28\pm 2.75$ & $3.95\pm 0.71$ & 25.4\\
\midrule
RefLDM                  & 3                      & $0.845\pm 0.063$ & $0.847\pm 0.045$ & $0.635\pm 0.092$ & $0.2117\pm 0.0574$ & $73.87\pm 3.92$ & $4.53\pm 0.63$ & 27.2\\
Ours                    & 3                      & $0.861\pm 0.067$ & $0.859\pm 0.049$ & $0.683\pm 0.077$ & $0.2040\pm 0.0602$ & $75.29\pm 2.75$ & $3.96\pm 0.71$ & 25.5\\
\midrule
RefLDM                  & 4                      & $0.848\pm 0.061$ & $0.848\pm 0.044$ & $0.639\pm 0.090$ & $0.2101\pm 0.0573$ & $74.26\pm 3.66$ & $4.50\pm 0.63$ & 27.2\\
Ours                    & 4                      & $0.863\pm 0.066$ & $0.859\pm 0.049$ & $0.680\pm 0.078$ & $0.2039\pm 0.0602$ & $75.29\pm 2.75$ & $3.96\pm 0.71$ & 25.5\\
\midrule
RefLDM                  & 5                      & $0.848\pm 0.060$ & $0.848\pm 0.043$ & $0.641\pm 0.090$ & $0.2097\pm 0.0574$ & $74.51\pm 3.52$ & $4.48\pm 0.64$ & 27.1\\
Ours                    & 5                      & $0.863\pm 0.066$ & $0.859\pm 0.048$ & $0.678\pm 0.079$ & $0.2038\pm 0.0601$ & $75.29\pm 2.75$ & $3.96\pm 0.71$ & 25.5\\
\bottomrule
\end{tabular}}
\end{table}

\begin{table}[h]
\caption{Detailed Quantitative Results on FFHQ-Ref Severe test set. Note, our multi-reference face support is training-free, while RefLDM's is not.}
\label{tab:severe-multi-full}
\resizebox{\linewidth}{!}{%
\begin{tabular}{lc|ccccccc}
\toprule
\multirow{2}{*}{Method} & \multirow{2}{*}{\#REF} & \multicolumn{7}{c}{FFHQ-Ref Severe}                 \\
\cmidrule{3-9}
                        &                        & IDS & FaceNet & IDS(REF) & LPIPS & MUSIQ & NIQE & FID \\
\midrule
RefLDM                  & 1 & $0.571\pm 0.110$ & $0.733\pm 0.052$ & $0.554\pm 0.112$ & $0.3366\pm 0.0756$ & $74.32\pm 3.36$ & $4.52\pm 0.62$ & 36.0\\
Ours                    & 1 & $0.609\pm 0.089$ & $0.743\pm 0.048$ & $0.712\pm 0.068$ & $0.3647\pm 0.0722$ & $75.22\pm 2.46$ & $3.84\pm 0.64$ & 38.3\\
\midrule
RefLDM                  & 2                      & $0.631\pm 0.091$ & $0.754\pm 0.049$ & $0.576\pm 0.100$ & $0.3271\pm 0.0745$ & $74.82\pm 3.20$ & $4.51\pm 0.62$ & 35.4\\
Ours                    & 2                      & $0.640\pm 0.078$ & $0.752\pm 0.047$ & $0.650\pm 0.073$ & $0.3625\pm 0.0717$ & $75.20\pm 2.42$ & $3.82\pm 0.63$ & 38.2\\
\midrule
RefLDM                  & 3                      & $0.662\pm 0.084$ & $0.764\pm 0.047$ & $0.594\pm 0.095$ & $0.3228\pm 0.0740$ & $75.22\pm 2.90$ & $4.49\pm 0.64$ & 35.1\\
Ours                    & 3                      & $0.652\pm 0.075$ & $0.755\pm 0.047$ & $0.636\pm 0.074$ & $0.3619\pm 0.0715$ & $75.19\pm 2.46$ & $3.82\pm 0.63$ & 38.4\\
\midrule
RefLDM                  & 4                      & $0.677\pm 0.080$ & $0.769\pm 0.047$ & $0.604\pm 0.093$ & $0.3203\pm 0.0731$ & $75.46\pm 2.73$ & $4.46\pm 0.64$ & 34.7\\
Ours                    & 4                      & $0.657\pm 0.074$ & $0.757\pm 0.048$ & $0.630\pm 0.075$ & $0.3617\pm 0.0715$ & $75.20\pm 2.42$ & $3.82\pm 0.63$ & 38.3\\
\midrule
RefLDM                  & 5                      & $0.685\pm 0.078$ & $0.772\pm 0.048$ & $0.611\pm 0.091$ & $0.3201\pm 0.0733$ & $75.62\pm 2.68$ & $4.46\pm 0.66$ & 34.7\\
Ours                    & 5                      & $0.658\pm 0.074$ & $0.757\pm 0.049$ & $0.626\pm 0.077$ & $0.3615\pm 0.0714$ & $75.20\pm 2.42$ & $3.83\pm 0.63$ & 38.2\\
\bottomrule
\end{tabular}}
\end{table}

\begin{table}[h]
\caption{Detailed Quantitative Results on CelebA-Ref-Test test set. Note, our multi-reference face support is training-free, while RefLDM's is not.}
\label{tab:celeba-multi-full}
\resizebox{\linewidth}{!}{%
\begin{tabular}{lc|ccccccc}
\toprule
\multirow{2}{*}{Method} & \multirow{2}{*}{\#REF} & \multicolumn{7}{c}{CelebA-Ref-Test}                 \\
\cmidrule{3-9}
                        &                        & IDS & FaceNet & IDS(REF) & LPIPS & MUSIQ & NIQE & FID \\
\midrule
RefLDM                  & 1 & $0.768\pm 0.085$ & $0.821\pm 0.046$ & $0.564\pm 0.096$ & $0.2453\pm 0.0550$ & $72.11\pm 4.59$ & $4.75\pm 0.55$ & 19.4\\
Ours                    & 1 & $0.779\pm 0.086$ & $0.827\pm 0.048$ & $0.691\pm 0.064$ & $0.2310\pm 0.0540$ & $75.64\pm 2.44$ & $3.98\pm 0.53$ & 18.4\\
\midrule
RefLDM                  & 2 & $0.775\pm 0.081$ & $0.824\pm 0.045$ & $0.580\pm 0.095$ & $0.2428\pm 0.0545$ & $73.01\pm 4.18$ & $4.69\pm 0.57$ & 18.8\\
Ours                    & 2 & $0.787\pm 0.084$ & $0.831\pm 0.047$ & $0.675\pm 0.071$ & $0.2305\pm 0.0540$ & $75.65\pm 2.43$ & $3.98\pm 0.53$ & 18.4\\
\midrule
RefLDM                  & 3 & $0.774\pm 0.080$ & $0.824\pm 0.044$ & $0.587\pm 0.095$ & $0.2426\pm 0.0542$ & $73.46\pm 4.02$ & $4.63\pm 0.56$ & 18.4\\
Ours                    & 3 & $0.787\pm 0.084$ & $0.831\pm 0.047$ & $0.668\pm 0.076$ & $0.2305\pm 0.0540$ & $75.65\pm 2.43$ & $3.98\pm 0.53$ & 18.4 \\
\midrule
RefLDM                  & 4 & $0.771\pm 0.080$ & $0.824\pm 0.044$ & $0.591\pm 0.095$ & $0.2434\pm 0.0542$ & $73.73\pm 3.93$ & $4.59\pm 0.57$ & 18.1\\
Ours                    & 4 & $0.786\pm 0.084$ & $0.831\pm 0.047$ & $0.664\pm 0.080$ & $0.2305\pm 0.0540$ & $75.65\pm 2.43$ & $3.98\pm 0.53$ & 18.4\\
\midrule
RefLDM                  & 5 & $0.767\pm 0.081$ & $0.822\pm 0.045$ & $0.594\pm 0.096$ & $0.2445\pm 0.0542$ & $73.93\pm 3.88$ & $4.56\pm 0.57$ & 18.0\\
Ours                    & 5 & $0.785\pm 0.085$ & $0.830\pm 0.046$ & $0.661\pm 0.082$ & $0.2306\pm 0.0540$ & $75.65\pm 2.43$ & $3.98\pm 0.53$ & 18.4\\
\bottomrule
\end{tabular}}
\end{table}

\begin{table}[h]
\caption{Ablation study on the classifier-free guidance scale parameters with FFHQ-Ref Severe.}
\label{tab:supp-cfg}
\resizebox{\linewidth}{!}{%
\begin{tabular}{lc|ccccccc}
\toprule
\multirow{2}{*}{$s_i$} & \multirow{2}{*}{$s_c$} & \multicolumn{7}{c}{FFHQ-Ref Severe}                 \\
\cmidrule{3-9}
                        &                        & IDS & FaceNet & IDS(REF) & LPIPS & MUSIQ & NIQE & FID \\
\midrule

1.0 & 1.0 & $0.599\pm 0.089$ & $0.738\pm 0.049$ & $0.694\pm 0.070$ & $0.3645\pm 0.0723$ & $74.73\pm 2.73$ & $3.97\pm 0.61$ & 39.1\\

1.0 & 1.2 & $0.608\pm 0.088$ & $0.742\pm 0.048$ & $0.719\pm 0.065$ & $0.3678\pm 0.0723$ & $75.13\pm 2.57$ & $3.94\pm 0.63$ & 38.8\\

1.2 & 1.0 & $0.598\pm 0.091$ & $0.738\pm 0.050$ & $0.685\pm 0.073$ & $0.3642\pm 0.0724$ & $74.84\pm 2.65$ & $3.85\pm 0.63$ & 38.8\\

1.2 & 1.2 & $0.609\pm 0.089$ & $0.743\pm 0.048$ & $0.712\pm 0.068$ & $0.3647\pm 0.0722$ & $75.22\pm 2.46$ & $3.84\pm 0.64$ & 38.3\\

\bottomrule
\end{tabular}}
\end{table}

\end{document}